\documentclass[conference]{IEEEtran}
\IEEEoverridecommandlockouts
\usepackage[caption=false,font=normalsize,labelfont=rm,textfont=rm]{subfig}
\usepackage{stfloats}
\usepackage{cite}
\newcommand{\fref}[1]{Figure \ref{#1}}

\newcommand{\tref}[1]{Table \ref{#1}}
\newcommand{\eref}[1]{Eq. (\ref{#1})}

\newcommand{\alref}[1]{Algorithm \ref{#1}}
\usepackage{graphicx}
\usepackage{amsmath}
\usepackage{amsthm}

\usepackage{amssymb,amsfonts}
\usepackage{algorithmicx}
\usepackage{array}
\usepackage{multirow}
\usepackage{mathrsfs}%
\usepackage{xcolor}%
\usepackage{textcomp}%
\usepackage{manyfoot}%
\usepackage{booktabs}%
\usepackage{algorithm}%
\usepackage{algpseudocode}%
\usepackage{listings}%
\usepackage{hyperref}
\usepackage{balance}
\usepackage{tabularx}
\usepackage{enumitem}
\setlist[enumerate]{label={(\arabic*)}}
\usepackage{soul}
\usepackage{array}
\usepackage{url}
\usepackage{verbatim}
\usepackage{epsfig}
\usepackage{color}
\usepackage{colortbl}
\usepackage{lineno}
\usepackage{graphics}
\usepackage{lmodern}
\graphicspath{{figures/}}
\usepackage{rotating}
\usepackage{float}
\usepackage{cases}
\usepackage{supertabular}
\usepackage{makecell}
\usepackage{url}
\usepackage{etoolbox} 

\RenewDocumentEnvironment{appendices}{}
{
    \renewcommand{\thesection}{\Alph{section}} 
    \setcounter{section}{0} 
    \section*{Appendices} 
    \addcontentsline{toc}{section}{Appendices} 
}
{}

\def\BibTeX{{\rm B\kern-.05em{\sc i\kern-.025em b}\kern-.08em
    T\kern-.1667em\lower.7ex\hbox{E}\kern-.125emX}}
\begin{document}

\title{PlatMetaX: An Integrated MATLAB platform for Meta-Black-Box Optimization}


\maketitle

\begin{abstract}
    The landscape of optimization problems has become increasingly complex, necessitating the development of advanced optimization techniques. Meta-Black-Box Optimization (MetaBBO), which involves refining the optimization algorithms themselves via meta-learning, has emerged as a promising approach. Recognizing the limitations in existing platforms, we presents PlatMetaX, a novel MATLAB platform for MetaBBO with reinforcement learning. PlatMetaX integrates the strengths of MetaBox and PlatEMO, offering a comprehensive framework for developing, evaluating, and comparing optimization algorithms. The platform is designed to handle a wide range of optimization problems, from single-objective to multi-objective, and is equipped with a rich set of baseline algorithms and evaluation metrics. We demonstrate the utility of PlatMetaX through extensive experiments and provide insights into its design and implementation. PlatMetaX is available at: \href{https://github.com/Yxxx616/PlatMetaX}{https://github.com/Yxxx616/PlatMetaX}.
\end{abstract}

\begin{IEEEkeywords}
evolutionary optimization, meta-black-box optimization, reinforcement learning, automated algorithm design, PlatEMO, MetaBox
\end{IEEEkeywords}

\section{Introduction}
\begin{figure}
    \centering
    \includegraphics[width=0.2\textwidth]{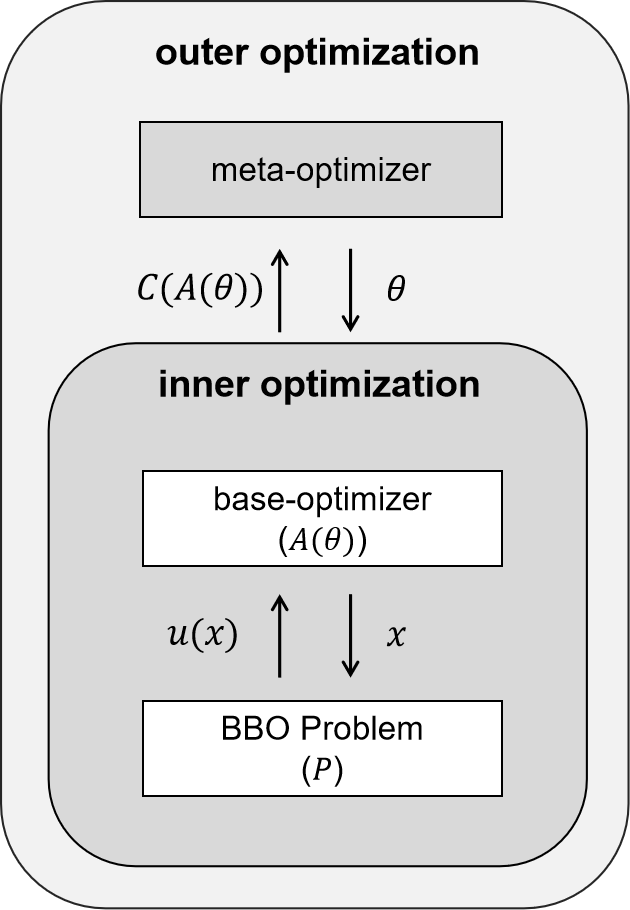}
    \caption{Bi-level optimization framework of MetaBBO.}
    \label{metabboframework}
\end{figure}

Optimization problems are fundamental to numerous fields, including engineering, economics, and artificial intelligence. As the complexity of these problems grows, traditional optimization methods often fall short, particularly when faced with high-dimensional, non-linear, multi-objective, or dynamic optimization landscapes \cite{zhou2024evolutionary, jiang2022evolutionary,zhan2022survey}. This has led to a surge in interest in meta-optimization, which involves optimizing the parameters of optimization algorithms to improve their performance across a variety of problems \cite{song2024reinforcement,li2019differential}. With the developing of meta-optimization, Meta-Black-Box Optimization (MetaBBO) emerges as a novel and effective framework, which employs meta-learning to design black-box optimizers automatically \cite{lange2022les}. It is a sophisticated framework that uses meta-learning to improve the effectiveness of traditional Black-box Optimization (BBO) methods, which inspired by the concept of "meta" to highlight its ability to adapt optimization strategies based on past experiences, much like how meta-learning algorithms adapt over time. It is done within a two-level optimization framework \fref{metabboframework} , where the optimizer used to optimize the black-box optimizers in the meta-level is termed meta-optimizer, while the black-box optimizers used to solve the specific problem is termed base-optimizer. 

With the developing of meta-learning, MetaBBO has been a powerful tool across diverse practical application domains robot control \cite{narayan2007intelligent,roveda2020robot}, healthcare and diagnosis \cite{waring2020automated,mustafa2021automated, an2024efficacy}, automated design of fan blades \cite{idahosa2008automated}, shop scheduling \cite{cheng2022multi,zhang2023q2}, vehicular ad-hoc networks \cite{nahar2023metalearn}, dynamic task allocation of crowdsensing \cite{ji2021q} and so on. Newly, Yang et al. have reviewed the MetaBBO for evolutionary algorithms (EAs) \cite{yang2025metareview}.

Despite their significant contributions to the field by providing environments for benchmarking and developing optimization algorithms, there remains a need for a unified platform that integrates the strengths of existing platforms and extends their capabilities for MetaBBO. 

PlatMetaX is introduced as a comprehensive platform that addresses these needs. It is designed to facilitate the development, evaluation, and comparison of MetaBBO methodologies within a unified environment. PlatMetaX leverages the capabilities of MATLAB, a widely and friendly used platform for numerical computing, to provide a robust framework.

\begin{figure*}
    \centering
    \includegraphics[width = 0.95\textwidth]{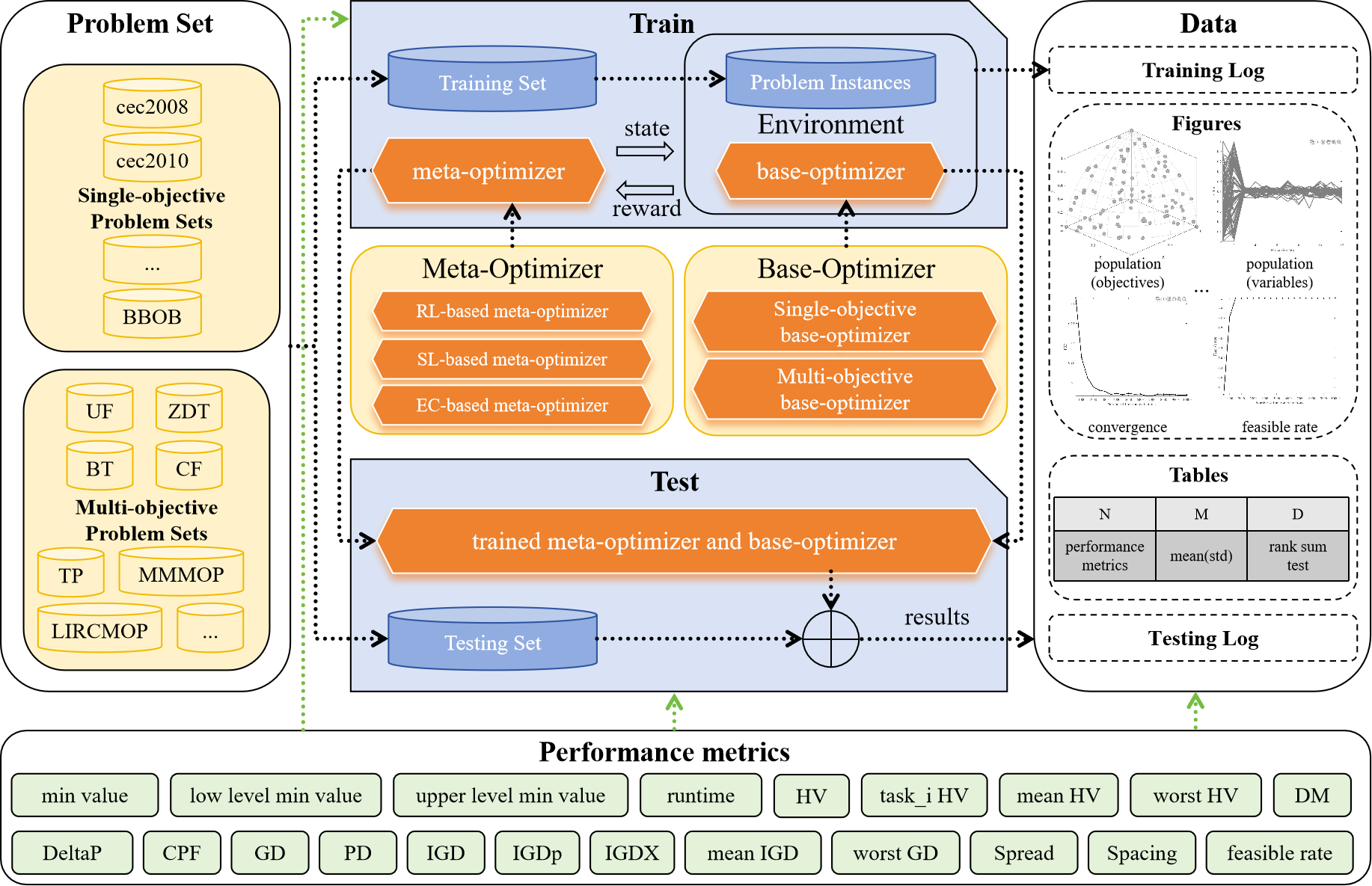}
    \caption{The architecture of PlatMetaX. }
    \label{PlatMetaBBOCons}
\end{figure*}

The architectural framework of PlatMetaX is illustrated in \fref{PlatMetaBBOCons}. This platform has been designed to support diverse meta-optimizers, incorporating a comprehensive suite of base-optimizers tailored for both single-objective and multi-objective optimization problems. This integration facilitates the exploration of complex optimization landscapes and enables the development of adaptive algorithms for various problem instances. A robust set of evaluation metrics is provided to ensure thorough performance assessment. The platform's design philosophy emphasizes modularity, user-friendliness, and the integration of state-of-the-art optimization and machine learning techniques. The following key contributions are highlighted:

\begin{itemize}
    \item MATLAB is utilized as the foundational platform for MetaBBO due to its robust capabilities in scientific computing and its user-friendly environment, which requires minimal configuration. This choice facilitates seamless integration of official tools for reinforcement learning (RL) and neural networks, enabling the development of meta-optimizers with ease.
    \item A suite of default meta-optimizers is provided, allowing users to select and deploy meta-optimizers without the need for parameter adjustments, beneficial from the design of PlatMetaX architectural. This feature is particularly beneficial for users whose focus is not on the development of meta-optimizers.
    \item Built upon the PlatEMO framework, PlatMetaX inherits a comprehensive library of algorithms and problems from PlatEMO, enriching the base library for designing base-optimizers and facilitating comparative studies between MetaBBO algorithms and traditional algorithms.
    \item Based on modifying the PlatEMO, PlatMetaX is equipped with a similar graphical user interface (GUI), enhancing the ease of experimental testing and the visualization of results.
    \item Two novel metrics are introduced to evaluate the transferability and generalization capabilities of the meta-optimizers, providing a more nuanced assessment of their performance.
    \item The workflow logic for meta-optimizers based on various methodologies, including RL, supervised learning (SL), and particularly evolutionary computation (EC), is standardized. This standardization enables users to rapidly develop and investigate diverse types of meta-optimizers within the PlatMetaX platform.
\end{itemize}

In summary, PlatMetaX represents a significant advancement in the field of optimization by providing a unified, comprehensive, and flexible platform for MetaBBO. Its capabilities are expected to greatly facilitate research and development in this area, leading to the discovery of more effective optimization algorithms capable of tackling the complex problems of the future.

\section{Preliminary}
Research in MetaBBO can be divided into two main areas based on the type of optimization targets:
\begin{itemize}
    \item \textbf{Parameter Configuration:} This area focuses on configuring the parameters of existing BBO algorithms. These parameters can be numerical (like mutation rates in genetic algorithms or learning rates in machine learning models) or structural (like initialization methods or network architectures). The goal here is to automatically find parameter settings that works well on the given problem offline, or dynamically control parameters online.

    \item \textbf{Strategy Generation:} This area goes beyond parameter configuration to create novel optimization strategies. By learning from a variety of problems, MetaBBO can develop new ways to solve optimization problems that might not be obvious using traditional methods.
\end{itemize}

\begin{figure}
    \centering
    \includegraphics[width=0.4\textwidth]{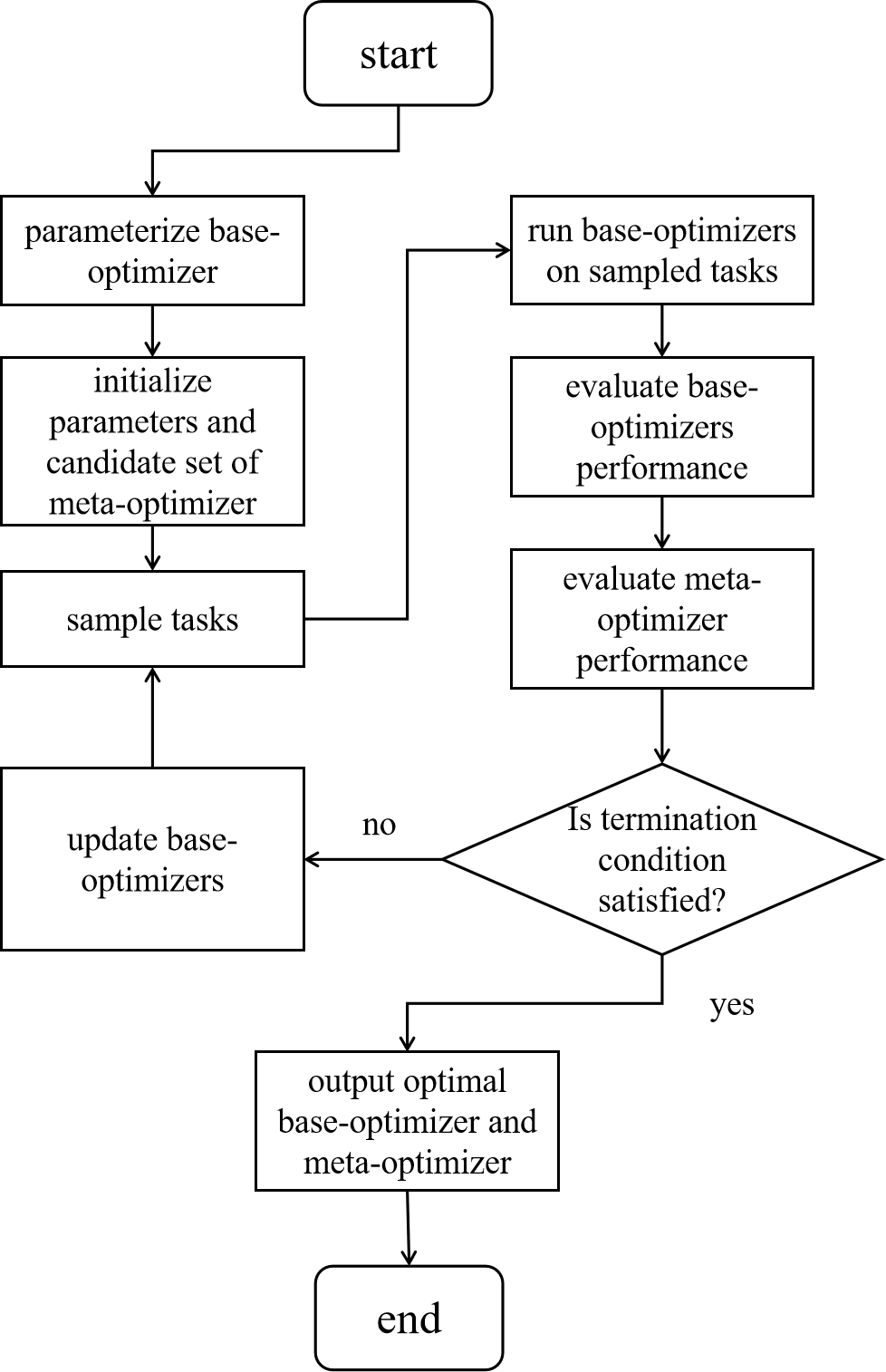}
    \caption{The flowchart of MetaBBO.}
    \label{figflowMetaBBO}
\end{figure}

The flowchart of MetaBBO is depicted in \fref{figflowMetaBBO}. During the learning process, the inner optimization (base-level) uses a base-optimizer to find the best solution for a given problem. The outer optimization (meta-level) uses a meta-optimizer to improve the base-optimizer's performance. Both levels generate a set of candidates: the outer level generates different base-optimizers, and the inner level generates solutions for the problem. This process helps in developing optimization methods that are not only effective but also generalizable across different problem types.

The first step in MetaBBO is to define the parameters of the base-optimizer. These parameters form a parameter space, denoted as $\Theta$. A BBO algorithm with a specific parameter configuration $\theta$ (where $\theta \in \Theta$) is represented as $A(\theta)$.

Next, the outer optimization's control parameters and candidate set are initialized. The control parameters include the termination condition for the outer loop and the meta-optimizer's settings. The candidate set is the set of configurations for the base-optimizer.

Then, a training set of BBO problems $P$ is selected from a distribution set $\mathcal{D}$, which covers different problem domains. The inner loop solves each problem in $P$ using $A(\theta)$. The performance of $A(\theta)$ is evaluated on $P$, and the meta-optimizer $\mathcal{C}(A(\theta))$ is defined as the expected performance of $A(\theta)$ across the true distribution of problems $\mathcal{D}$:

\begin{equation}
\mathcal{C}(A(\theta)) = \mathbb{E}_{p \in P, A \subseteq P, A \sim d, d \in \mathcal{D}}[\mathcal{J}(A(\theta), p)]
\end{equation}
where $\mathcal{J}(A(\theta), p)$ is the performance of $A(\theta)$ on problem $p$.

The outer loop updates the base-optimizers, samples new problems $P$, runs the base-optimizers, and evaluates the meta-optimizer until a stopping condition is met.

The optimization goal in MetaBBO is to find the best parameter configuration $\theta^*$ that maximizes the expected performance:

\begin{equation}
\theta^* \in \arg\max_{\theta \in \Theta} \mathbb{E}_{p \in P, A \subseteq P, A \sim d, d \in \mathcal{D}}[\mathcal{J}(A(\theta), p)]
\end{equation}

\textbf{Related benchmarks.} MetaBox \cite{ma2024metabox} and PlatEMO \cite{tian2017platemo} are notable platforms that have significantly contributed to the field of optimization by providing comprehensive environments for benchmarking and developing intelligent computation. MetaBox is the first benchmark platform for MetaBBO with RL (MetaBBO-RL), simplifying the development of MetaBBO-RL and ensure an automated workflow with Python. PlatEMO, is a MATLAB-based platform for evolutionary optimization, featuring a wide array of traditional evolutionary algorithms and problems. Newest PlatEMO includes nearly 300 evolutionary algorithms and more than 500 test problems, along with several widely used performance indicators. With a user-friendly graphical user interface, PlatEMO enables users to easily compare several evolutionary algorithms at one time and collect statistical results in Excel or LaTeX files.

Both platforms have facilitated extensive research and development in optimization. However, MetaBox focus on RL-assisted MetaBBO and single-objective optimization, and the baselines are limited. PlatEMO has not encountered with MetaBBO. There is a clear need for a platform that can integrate meta-optimization techniques with a broad spectrum of base-optimizers, catering to both single-objective and multi-objective problems. 

PlatMetaX is designed to fill this gap by providing a unified environment that supports the development and evaluation of meta-optimization algorithms using RL, neural networks, and evolutionary computation. This platform aims to enhance the adaptability and efficiency of optimization algorithms, making them more effective in tackling the diverse and challenging problems encountered in various fields. 

The comparison between related benchmarks are shown in \tref{tabComp}.

\begin{table*}[htbp]
    \centering
    \caption{Comparison between related benchmarks}
      \begin{tabular}{cccccccccc}
      \toprule
      \multirow{2}[2]{*}{Platform} & \multirow{2}[2]{*}{Language} & \multicolumn{1}{c}{\multirow{2}[2]{*}{Problem Number}} & \multicolumn{2}{c}{Problem Type} & \multirow{2}[2]{*}{Baseline} & \multirow{2}[2]{*}{GUI} & \multicolumn{3}{c}{Meta-support} \\
            &       &       & SOO   & MOO   &       &       & RL-support & SL-support & EC-support \\
      \midrule
      MetaBox & PYTHON & 334   & \checkmark & $\times$ & 19    & $\times$ & \checkmark & $\times$ & $\times$ \\
      PlatEMO(v4.11) & MATLAB & 583   & \checkmark & \checkmark & 302   & \checkmark & $\times$ & $\times$ & $\times$ \\
      PlatMetaX & MATLAB & 583   & \checkmark & \checkmark & 302+  & \checkmark & \checkmark & \checkmark & \checkmark \\
      \bottomrule
      \end{tabular}%
    \label{tabComp}%
\end{table*}%

\section{PlatMetaX: Design and Resources}
PlatMetaX is structured around a modular design that facilitates the integration of various meta-optimizers and base-optimizers. 

\subsection{Templates}
\begin{figure*}
    \centering
    \includegraphics[width=0.8\textwidth]{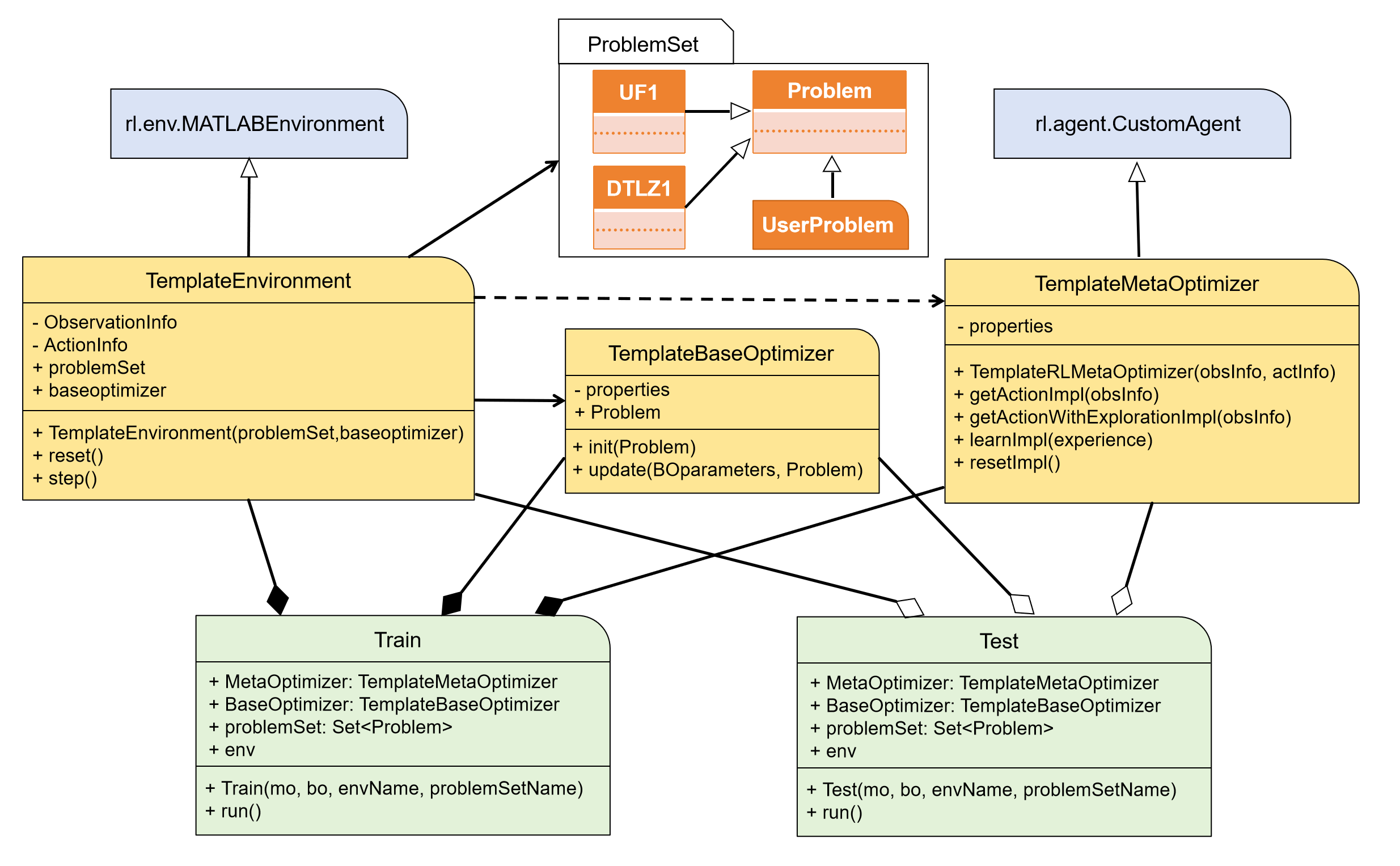}
    \caption{UML class of PlatMetaX. Users can define own methods from provided templates highlighted in yellow to enable polymorphism. Classes in blue box are the official classes of MATLAB, while classes in orange box are the original classes of PlatEMO.}
    \label{figtemplate}
\end{figure*}

PlatMetaX is modular and equipped with flexible templates intended for users developing MetaBBO algorithms. These templates are designed to be extensible, enabling the straightforward integration of new ideas and techniques. A UML class diagram, as shown in Figure \ref{figtemplate}, provides a clear representation of the templates and the relationships between key classes. An overview of these classes is provided as follows:

\begin{itemize}
\item \texttt{TemplateEnvironment}: Constitutes the foundational environment framework, which is responsible for defining the observation and action spaces. It orchestrates the problem-feeding mechanism as well.
\item \texttt{TemplateBaseOptimizer}: Represents the foundational class for base-optimizer, which is waiting to be refined by meta-optimizer.
\item \texttt{TemplateMetaOptimizer}: Inherits from \texttt{rl.agent.CustomAgent} and generates specific improvements (collectively referred to as action) for base-optimizer.
\item \texttt{Problem}: Serves as the optimization instances for training or testing, originating from PlatEMO. Users can facilitate the creation of new optimization problems by inheriting this class and \texttt{UserProblem} provides a template. 
\item \texttt{Train} and \texttt{Test}: Manage the training and testing phases of the MetaBBO workflow, respectively.
\end{itemize}

\subsubsection{TemplateEnvironment}
\texttt{TemplateEnvironment}, which is derived from \texttt{rl.env.MATLABEnvironment}. This class serves as a fundamental structure for defining custom environments within MATLAB, tailored for optimization tasks. It encapsulates the necessary properties and methods to manage the environment's interactions with the meta- and base-optimizers. 
possible
Several properties can be defined as follows:
\begin{itemize}
\item \texttt{curProblem}: Represents the current problem being optimized.
\item \texttt{baseoptimizer}: Refers to the base-optimizer instance being used.
\item \texttt{problemSet}: Contains a set of problems to be optimized.
\item \texttt{State}: Represents the situation that the meta-optimizer faces.
\item \texttt{IsDone}: A protected property indicating whether the episode has terminated.
\end{itemize}

The constructor \texttt{TemplateEnvironment} initializes a new instance of the environment. It defines the observation and action specifications, and configures the problems and the base-optimizer via the incoming parameters the to be optimized during training phase or testing phase. Several methods need to be implemented to manage the environment's operations:
\begin{itemize}
\item \texttt{reset}: Resets the environment to its initial state by calculaing current problem features and initializing the base-optimizer. It returns the initial observation based on whether the problem is single-objective or multi-objective.
\item \texttt{step}: Takes an action and updates the environment's state. It returns the observation, reward, a flag indicating whether the episode is done, and logged signals. This method also updates the best population and saves test results if necessary.
\end{itemize}

\subsubsection{TemplateBaseOptimizer}
This class is responsible for handling the initialization (\texttt{init}) and updating (\texttt{update}) of base optimizers in response to problem-specific parameters. On the meanwhile, the problem or population characteristics calculations are suggested be defined in this class.
The PlatMetaX platform offers a \texttt{TemplateBaseOptimizer} class designed for extensibility, enabling users to design custom base-optimizers. This template class is derived from \texttt{BASEOPTIMIZER} and is compatible with the functional components of PlatEMO, facilitating the direct utilization of its algorithmic capabilities. It encapsulates the properties and methods necessary for solving problem instances.
The key methods are initialization and update.
\begin{itemize}
\item \texttt{init}: Initializes the base-optimizer with a given problem. It sets up the population using the problem's initialization method, determines the number of constraints, and initializes other properties such as population size and hypervolume value. 
\item \texttt{update}: Updates the evolution state when solving the current problem using the base-optimizer configured by the incoming parameters. It usually performs selection, offspring generation through genetic operators, and environmental selection to update the population. 
\end{itemize}

\subsubsection{TemplateMetaOptimizer}
\begin{figure*}
    \centering
    \includegraphics[width=0.7\textwidth]{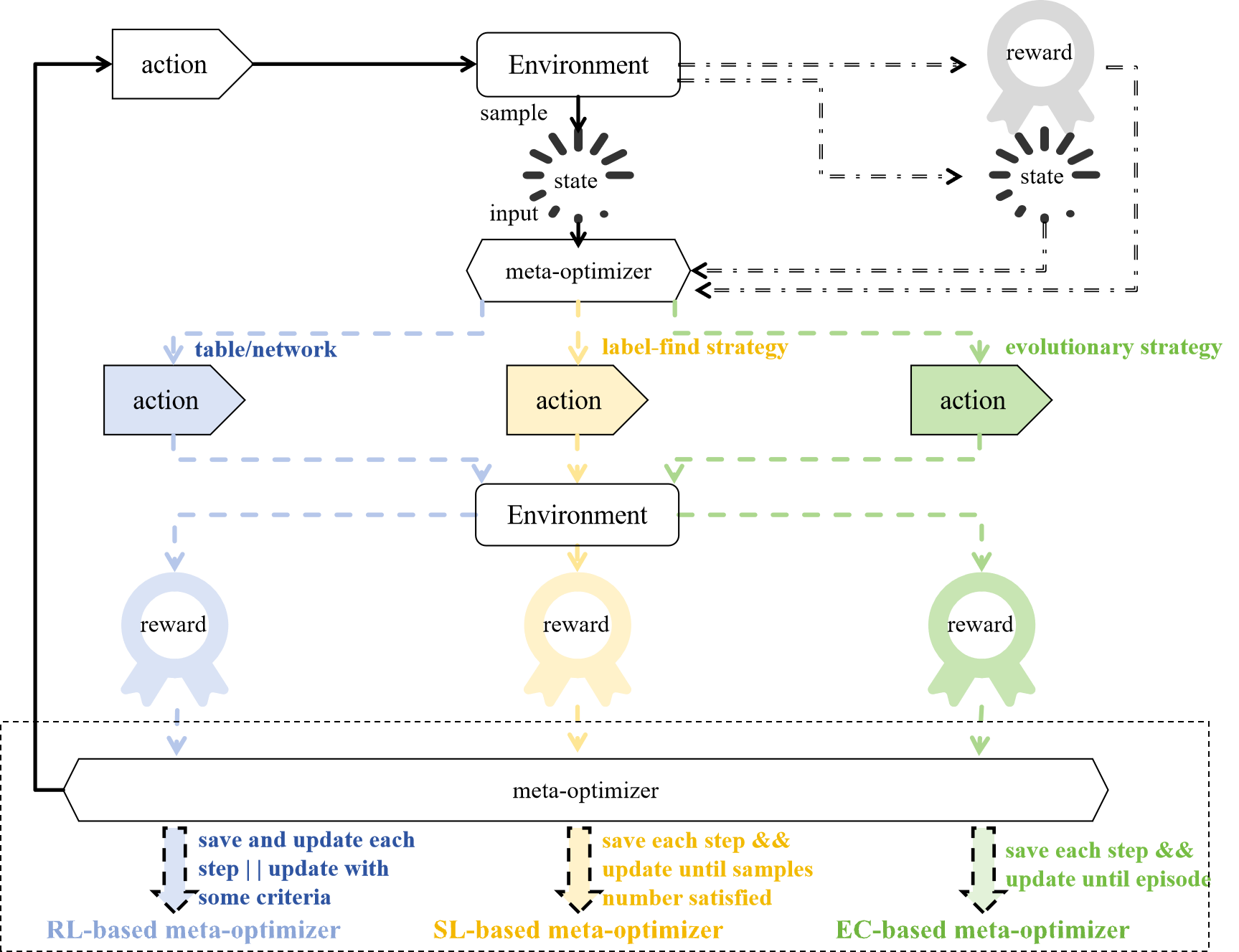}
    \caption{Unified working logic of three kinds of meta-optimizers, where the part in dashed box represents one step of the meta-optimizer. Different meta-optimizers can be implemented via the action generation strategies and specific step method which update the mapping policy.}
    \label{figworkinglogic}
\end{figure*}

Referring to the principle of RL, a unified working logic is introduced to adapt to various kinds of meta-optimizers. \fref{figworkinglogic} presented the unified working flow and the difference among three kinds of meta-optimizer.

Based on the working logic, a unified meta-optimizer template \texttt{TemplateMetaOptimizer} is provided, which is derived from MATLAB official document \texttt{rl.agent.CustomAgent}. It encompasses several properties that are pivotal for its operation. The constructor \texttt{TemplateRLMetaOptimizer} is designed to initialize the meta-optimizer with observation and action information. It defines the mapping policy. The method \texttt{getActionWithExplorationImpl} implements the action generation strategy during training, i.e. selecting actions by forwarding the observation through the mapping policy. Exploration noise, such as Gaussian noise, can be added to the action to encourage exploration for RL-based meta-optimizer, while it can be regarded as offspring generation if the meta-optimizer is EC-based, and it can be regarded as data-collecting and label-finding for SL-based meta-optimizer.  The method \texttt{getActionImpl} defines the action generation strategy during testing, which will generate fixed action. The \texttt{learnImpl} method realizes the mapping policy updating logic. Methods \texttt{getActionImpl},\texttt{getActionWithExplorationImpl} and \texttt{learnImpl} must be required to be implemented. In addition, the method \texttt{resetImpl} resets the agent's state, including clearing the experience buffer and resetting the network state if necessary. 

\subsection{Workflow}
\begin{figure*}
    \centering
    \subfloat[training phase \label{figtrain}]{\includegraphics[width = 0.8\textwidth]{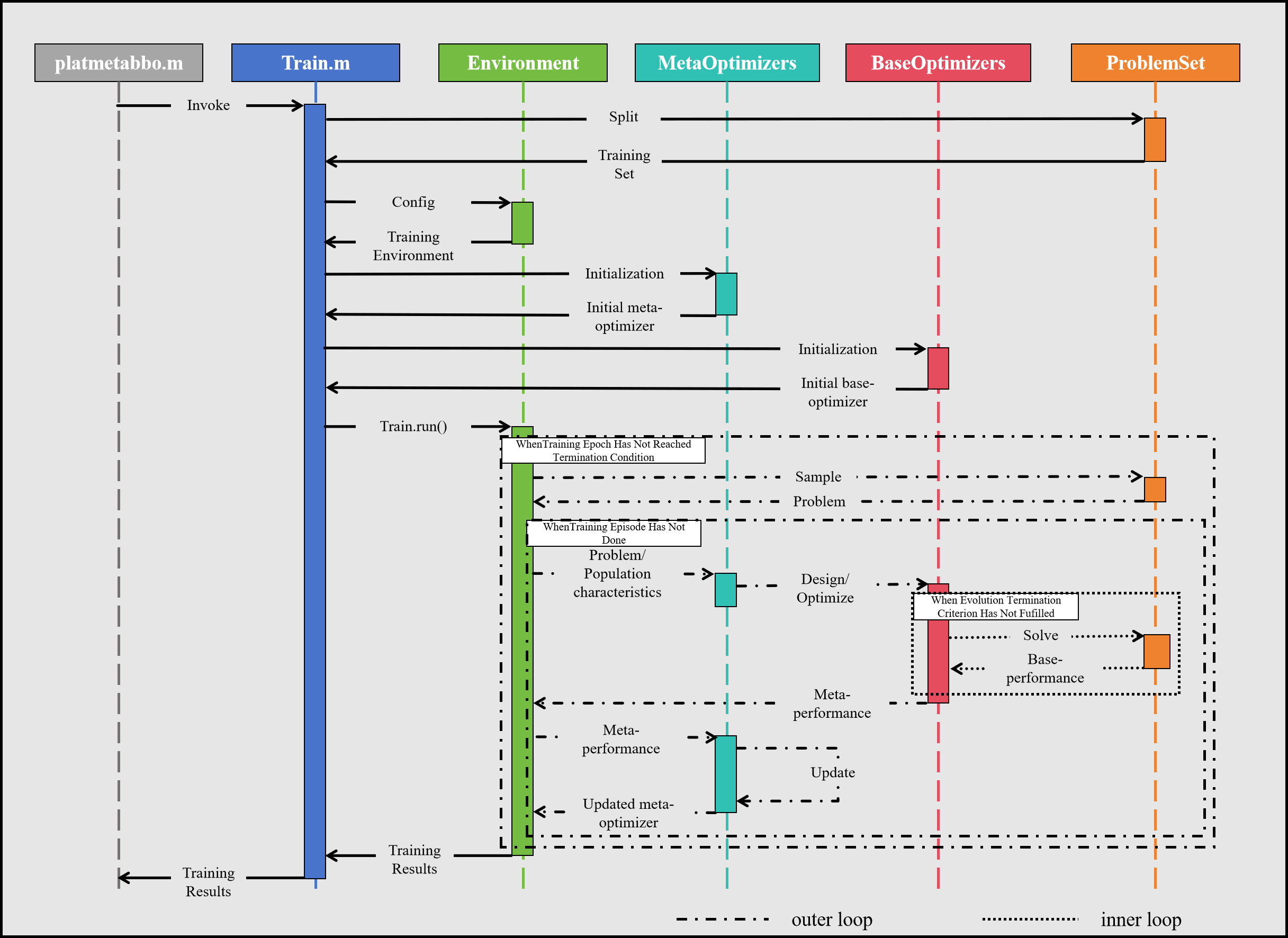}} \\
    \subfloat[testing phase \label{figtest}]{\includegraphics[width = 0.8\textwidth]{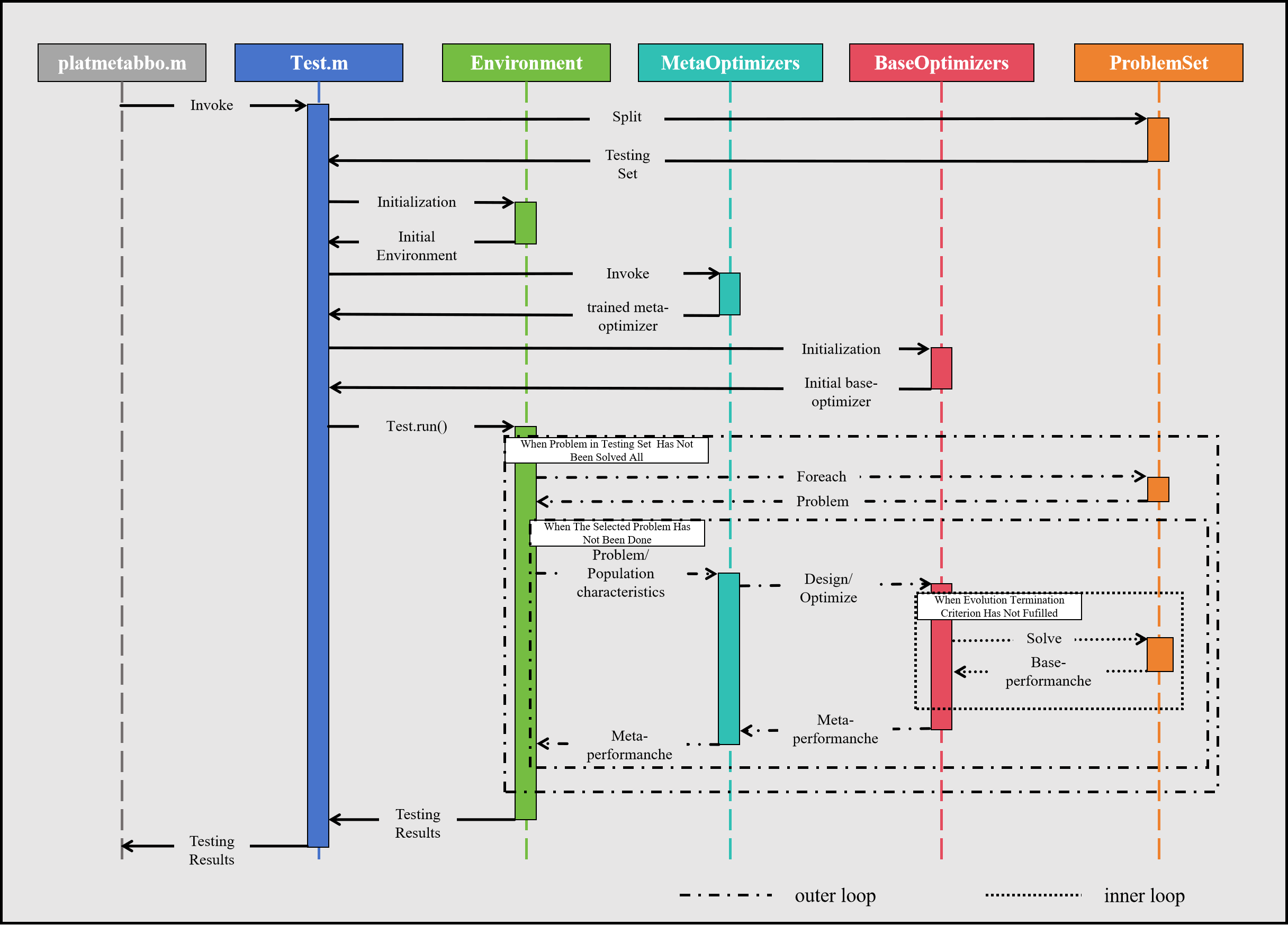}}
    \caption{The workflow of PlatMetaX.}
    \label{figPlatMetaBBO}
\end{figure*}

The workflow of PlatMetaX is meticulously designed to streamline the process of developing MetaBBO methodologies. As depicted in \fref{figPlatMetaBBO}, the workflow is bifurcated into training phase (\fref{figtrain}) and testing phase (\fref{figtest}), each serving distinct purposes in the lifecircle of MetaBBO.

The training phase is initiated by defining a training problem set, current task environment, the equipped meta-optimizer and base-optimizer. The environment plays a crucial role in this phase by sampling problem instance, informing the current situation to meta-optimizer, providing feedback in the form of performance of meta-optimizer (meta-performance), which are calculated based on the performance metrics of the base-optimizer (base-performance). The meta-optimizer in training phase is tasked with learning to improve the performance of the parameterized base-optimizer across a variety of problem instances. A feedback outer loop is essential for guiding the meta-optimizer's learning process, enabling it to converge towards an optimal set of parameters for the base-optimizer. The optimized base-optimizer is utilized to solve the sampled problem instances inner-iteratively and return base-performance.

The pseudo-code of the training phase in PlatMetaX is outlined as \alref{algTrain}.
\begin{algorithm*}
    \caption{Pseudo-code for Training}
    \begin{algorithmic}[1]
    \Require User-defined \texttt{MetaOptimizer} $M$, \texttt{BaseOptimizer} $B$, equipped environment name \texttt{envName}, user-specified problem set name \texttt{ProblemSet}
    \Ensure Trained \texttt{MetaOptimizer} and training information
    \State $trainProblemSet \gets$ specify training set from $ProblemSet$
    \State Initialize environment $env \gets envName(trainProblemSet, B, 'train')$
    \State Initialize $MetaOptimizer \gets M(env)$
    \State Initialize $BaseOptimizer \gets B()$
    \State Set training parameters $trainOpts$
    \State Train $MetaOptimizer$ using official command: $trainedMetaOptimizer \gets train(MetaOptimizer, env, trainOpts)$
    \State Save trained $MetaOptimizer$
    \State \Return training information
    \end{algorithmic}
    \label{algTrain}
\end{algorithm*}

Post the training phase, the meta-optimizer and the base-optimizer are subjected to a testing phase to evaluate their performance on unseen problem instances. This phase involves defining a separate testing set, ensuring that the evaluation is conducted on problems that were not included in the training phase. The results obtained from this phase provide insights into the generalizability and robustness of the developed algorithm. 

\subsection{Problem Sets}
A critical component of PlatMetaX is the diverse collection of problem sets that serve as benchmarks for evaluating the performance of optimization algorithms. This part directly utilized the problems from PlatEMO. These problem sets are meticulously curated to encompass a wide range of optimization challenges, ensuring that the algorithms developed using PlatMetaX are tested against both traditional and contemporary optimization landscapes. The problems are classified into two sets: Single-Objective Problems Set and Multi-Objective Problems Set.

\subsubsection{Single-Objective Problems Set}

These problems are designed to test the ability of optimization algorithms to converge to the global optimum in scenarios where a single objective function is to be minimized or maximized. The problems vary in complexity, including unimodal and multimodal functions, noiseless and noisy environments, and different search space dimensions. 

This problem set includes basic SOPs (such as Sphere, Schwefel, Rosenbrock, Step, Rastrigin, Ackley, Griewank, etc. \cite{yao1999evolutionary}), and benchmarks from CEC (Congress on Evolutionary Computation) competitions held in 2008 \cite{tang2007benchmark}, 2010 \cite{mallipeddi2010problem}, 2013 \cite{li2013benchmark}, 2017 \cite{wu2017problem}, and 2020 \cite{yue2020problem}. BBOB2009 (black-box optimization benchmarking 2009) \cite{hansen2009real} is introduced as well. Besides, there are some real-world SOPs (such as TSP \cite{corne2007techniques}, max-cut problem\cite{tian2024neural}, knapsack problem \cite{zitzler1999multiobjective}, and community detection problem with label based encoding \cite{tian2019evolutionary}), multitasking problem \cite{bali2019multifactorial}, and ISCSO (International student competition in structural optimization) \cite{azad2023standard}.

\subsubsection{Multi-Objective Problem Sets}

To address the growing need for multi-objective optimization, PlatMetaX integrates several multi-objective problem sets. These problems are crafted to evaluate the algorithms' capability to navigate the Pareto front, which represents the set of optimal trade-offs between multiple conflicting objectives.

This set includes numerous classic MOPs (such as BT (Benchmark MOP with bias feature) \cite{li2016biased}, CF (Constrained benchmark MOP) \cite{zhang2009multiobjective}, DAS-CMOP (Difficulty-adjustable and scalable constrained benchmark MOP) \cite{fan2020difficulty}, DTLZ \cite{deb2005scalable}, etc.), multitasking MOPs \cite{gupta2016multiobjective}, and MOPs with variable linkages \cite{zhang2008rm, cheng2015multiobjective,li2008multiobjective,liu2013decomposition}. Benchmark EvoXBench \cite{lu2023neural} from EvoX \cite{huang2024evox} is also introduced, which focuses on neural architecture search. Some real-world MOPs including multi-objective TSP \cite{corne2007techniques}, multi-objective next release problem \cite{zhang2007multi}, multi-objective quadratic assignment problem \cite{knowles2003instance}, multi-line distance minimization problem \cite{li2017multiline}, and sparse optimization problems \cite{tian2019sparse, tian2020solving,su2022comparing } also can be used for testing.

\subsection{Optimizers and Environment}
The core functionality of PlatMetaX is facilitated through the integration of meta-optimizers and base-optimizers within a specially designed environment. This section elaborates on the roles and characteristics of these optimizers and the environment that supports their operation.

\subsubsection{Meta-Optimizers}
Meta-optimizers in PlatMetaX are algorithms designed to refine the base-optimizers. They are dedicated to adapting the base-optimizers to perform effectively across a spectrum of problems. The platform supports three types of meta-optimizers:
\begin{itemize}
\item \textbf{RL-based meta-optimizers} are algorithms that learn to optimize the parameters of base-optimizers through trial and error, guided by a reward signal that reflects the performance of the base-optimizer \cite{LU2025101783, 10496708,10144924,10049395,LIAO2024101568,LUO2025121648,9852781,chayboutihal03930140}. 
\item \textbf{SL-based meta-optimizers} utilize supervised learning to predict and adjust the parameters of base-optimizers, leveraging the ability of neural networks to model complex relationships \cite{9187549,GUTIERREZRODRIGUEZ2019470,oyelade2025deep,chen2017learning,pmlr-v202-krishnamoorthy23a}.
\item \textbf{EC-based meta-optimizers} employ evolutionary algorithms to evolve the parameters of base-optimizers, capitalizing on the global search capabilities of evolutionary computation \cite{10884874,lange2023discovering}.
\end{itemize}

These meta-optimizers are designed to be flexible and adaptable, enabling them to tailor the behavior of base-optimizers to the specific characteristics of the optimization problems at hand. 

We have implemented these three kinds of meta-optimizers as baselined. For RL-based meta-optimizer, DDPG\_DE\_F and DQN\_DE\_MS are implemented, which utilize DDPG and DQN to refining DE and work for continuous and discrete outputs separately. Specifically, DDPG\_DE\_F is designed to adjust scale factor $F$, while DQN\_DE\_MS is designed to select mutation strategies during evolution. For SL-based meta-optimizer, a MLP is designed to recommend single-objective optimization evolutionary algorithms including ABC, CSO, DE, PSO and SA for different TSPs, named MLP\_Alg\_Rec\_. For EC-based meta-optimizer, DE is used at meta-level to optimize the scale factor and cross rate of another DE at base-level, named DE\_DE\_FCR. 

In addition, with the developing of large language model (LLM), it has been an emerging tool for MetaBBO \cite{romera2024mathematical,zhong2025llmoa,zhong2024geminide,liu2024large}. Regarding this research, it will be added to our platform in future versions.

\subsubsection{Base-Optimizers}
Base-optimizers are the fundamental optimization algorithms that are subject to optimization by the meta-optimizers. PlatMetaX includes a variety of base-optimizers to cover different optimization scenarios:
\begin{itemize}
\item \textbf{Single-objective base-optimizers} are designed to minimize or maximize a single objective function. They include traditional algorithms such as Particle Swarm Optimization (PSO), Differential Evolution (DE), and Genetic Algorithms (GA). 
\item \textbf{Multi-objective base-optimizers} are capable of handling multiple conflicting objectives simultaneously. They include Pareto-based, indicator-based and decomposition based multi-objective evolutionary algorithms (MOEAs).
\end{itemize}

All the algorithms in the PlatEMO can be transformed to the base-optimizers via TemplateBaseOptimizer. Users only need to take the parameterization of the algorithm into consideration \cite{yang2025metareview}.

\subsubsection{Environment}
The environment in PlatMetaX is a critical component that interacts with both the meta-optimizer and the base-optimizer. It serves several key functions:
\begin{itemize}
\item \textbf{Problem Instance Management} is handled by the environment, which provides the base-optimizer with access to the problem instances selected from the problem sets.
\item \textbf{Performance Assessment} is conducted by the environment, which evaluates the performance of the base-optimizer using predefined metrics and provides feedback in the form of rewards to the meta-optimizer.
\item \textbf{State and Reward Interaction} is facilitated by the environment, which maintains the state of the optimization process and communicates the reward to the meta-optimizer based on the base-optimizer's performance.
\end{itemize}

\subsection{Performance Metrics}
Performance metrics are integrated to the evaluation process within PlatMetaX, providing a quantitative measure of the efficacy of both meta-optimizers and base-optimizers. These metrics are designed to offer a comprehensive assessment of the optimization algorithms' ability to navigate complex problem landscapes and achieve optimal solutions.

For single-objective optimization problems, the solution quality is used to assess the base-optimizers if the problem is unconstrained. While the problem is constrained, feasiblity rate is another evaluator.

For multi-objective optimization problems, a suite of optimization performance metrics is employed to evaluate the performance of base-optimizers, such as:
\begin{itemize}
\item \textbf{Inverted Generational Distance (IGD)} is utilized to quantify the distance between the true Pareto front and the approximate Pareto front obtained by the optimizer. A lower IGD value indicates a closer approximation to the true Pareto front.
\item \textbf{Generational Distance (GD)} measures the average distance between the solutions obtained by the optimizer and the true Pareto front. It is an indicator of how well the optimizer's solutions are spread out across the Pareto front.
\item \textbf{Spacing (Spacing)} assesses the uniformity of the solutions along the Pareto front. A uniform distribution of solutions is indicative of a well-performing optimizer.
\item \textbf{Hypervolume (HV)} is employed to measure the volume of the objective space dominated by the solutions obtained by the optimizer. A larger hypervolume indicates a more effective optimization.
\end{itemize}

The assessment of meta-optimizers is conducted through a set of straightforward yet informative metrics that focus on the overall optimization performance of the base-optimizer across solved problems within a single epoch. These metrics are designed to provide a clear and concise evaluation of the meta-optimizer's effectiveness in enhancing the base-optimizer's performance. The \textbf{average performance} of the base-optimizer is calculated by aggregating the performance metrics of the base-optimizer over all training problems for a given epoch. The \textbf{best performance} metric identifies the highest performance achieved by the base-optimizer on any single training problem within an epoch. This metric highlights the meta-optimizer's potential to significantly enhance the performance of the base-optimizer on particularly well-suited problems. The \textbf{worst performance} metric evaluates the lowest performance of the base-optimizer across all training problems in an epoch. 
These metrics collectively provide a balanced assessment of the meta-optimizer's performance, covering both its generalization and robustness. By monitoring these metrics throughout the training process, users can gain insights into how effectively the meta-optimizer is adapting the base-optimizer to handle the training problems.

PlatMetaX also supports the definition and implementation of custom metrics. This flexibility allows users to tailor the evaluation process to the specific requirements of their research, enabling a more targeted assessment of optimizer performance. 

More evaluation metrics of meta-optimizer will be introduced in the future work, including the comprehensive performance such as AEI \cite{ma2024metabox}, generalization performance and transfer performance. 

\section{Benchmarking Study}
PlatMetaX serves as a convenient tool for conducting experimental studies, allowing users to develop their own MetaBBO algorithms quickly, tune traditional evolutionary algorithms flexibly, comparing different algorithms intuitively. Several experimental examples including developing MetaBBO with DE as the base-optimizer and comparing the experimental results. 

\subsection{Experimental Setup}
Users need to train the meta-optimizer with indicated parameters including problem set (necessary) and its problem dimensions and problem objective numbers (unnecessary). The training options can to be specified including maximum episodes, maximum steps per episode and stopping criteria in class \texttt{Train}. In our experiments, BBOB2009 \cite{hansen2009real} is leveraged as the training and testing problem set, where the functions except $F_1,F_5,F_6,F_{10},F_{15},F_{20}$ are set as as training problems (referred to the easy-train mode in \cite{ma2024metabox}). In the training episode, the maximum steps is set to 1000, maximum steps per episode is set to 500, while the stopping criteria is set to the situation that when the average reward within 10-episode length slide window equals or exceed to a threshold 100. In the testing episode, each testing instance-algorithm experiment is conducted 31 independent replications and Wilcoxon-rank-sun test is utilized to assess the statistical significance. All results presented are obtained using a machine of Intel i7-10510U CPU with 16GB RAM. 

\subsection{Comparison of Different MetaBBO Algorithms and Traditional Algorithms}
We attempt to train two kinds of MetaBBO algorithms (DDPG\_DE\_F, DQN\_DE\_MS) as baselines on PlatMetaX. They are compared with traditional DE algorithm. The experimental results presented in Table~\ref{tabresD10} demonstrate the performance of DDPG\_DE\_F and DQN\_DE\_MS, compared to the traditional DE algorithm on the BBOB2009 benchmark suite where the problem dimension is 10. The results are analyzed based on the average and standard deviation of the optimization outcomes, with statistical significance indicated by the symbols "-", "+", and "=", representing significantly worse, significantly better, or equal performance compared to DE, respectively. 

\begin{table*}[htbp]
    \centering
    \caption{Per-instance optimization results $v_{avg}(v_{std})$ of testing instances on BBOB2009 with $D=10$, where the symbols "-","+", and "=" separately indicate whether the given algorithm performs significantly worse, significantly better, or equal to traditional DE.}
      \begin{tabular}{cccc}
      \toprule
      \textbf{Problem} & \textbf{DDPG\_DE\_F} & \textbf{DQN\_DE\_MS} & \textbf{DE} \\
      \midrule
      BBOB\_F1 & 2.9646e+1 (6.19e+0) - & \textcolor[rgb]{ .2,  .2,  .914}{1.2111e-4 (3.65e-4) +} & 1.6059e-1 (4.79e-2) \\
      BBOB\_F2 & 4.3404e+4 (1.67e+4) - & \textcolor[rgb]{ .2,  .2,  .914}{1.0476e+2 (2.51e+2) +} & 2.8781e+2 (1.06e+2) \\
      BBOB\_F3 & 1.8532e+2 (2.63e+1) - & \textcolor[rgb]{ .2,  .2,  .914}{1.7328e+1 (9.08e+0) +} & 5.1737e+1 (7.22e+0) \\
      BBOB\_F4 & 2.5790e+2 (4.01e+1) - & \textcolor[rgb]{ .2,  .2,  .914}{1.4722e+1 (8.50e+0) +} & 6.9303e+1 (7.11e+0) \\
      BBOB\_F5 & \textcolor[rgb]{ .2,  .2,  .914}{0.0000e+0 (0.00e+0) +} & 1.2662e+1 (5.64e+0) - & 6.5483e-1 (8.71e-1) \\
      BBOB\_F6 & 5.4591e+1 (1.13e+1) - & \textcolor[rgb]{ .2,  .2,  .914}{2.0110e-3 (7.80e-3) +} & 6.2501e-1 (2.01e-1) \\
      BBOB\_F7 & 9.7479e+1 (2.23e+1) - & \textcolor[rgb]{ .2,  .2,  .914}{5.7269e-2 (1.03e-1) +} & 9.3528e-1 (3.21e-1) \\
      BBOB\_F8 & 1.7180e+4 (6.18e+3) - & \textcolor[rgb]{ .2,  .2,  .914}{7.5866e+0 (5.11e+0) +} & 5.9440e+1 (1.92e+1) \\
      BBOB\_F9 & 1.9429e+3 (4.63e+2) - & \textcolor[rgb]{ .2,  .2,  .914}{1.1678e+1 (1.73e+1) +} & 6.4550e+1 (1.79e+1) \\
      BBOB\_F10 & 3.9750e+4 (2.31e+4) - & \textcolor[rgb]{ .2,  .2,  .914}{5.6883e+1 (7.59e+1) +} & 2.1641e+3 (8.19e+2) \\
      BBOB\_F11 & 1.4295e+2 (4.44e+1) - & \textcolor[rgb]{ .2,  .2,  .914}{1.5321e-1 (2.65e-1) +} & 1.3506e+1 (3.55e+0) \\
      BBOB\_F12 & 7.9995e+6 (1.85e+6) - & \textcolor[rgb]{ .2,  .2,  .914}{8.7039e+0 (1.59e+1) +} & 1.4381e+5 (5.58e+4) \\
      BBOB\_F13 & 8.4729e+2 (9.58e+1) - & \textcolor[rgb]{ .2,  .2,  .914}{1.9006e+0 (3.78e+0) +} & 6.7560e+1 (1.39e+1) \\
      BBOB\_F14 & 8.7194e+0 (2.53e+0) - & \textcolor[rgb]{ .2,  .2,  .914}{8.0060e-4 (2.62e-3) +} & 1.9796e-1 (6.17e-2) \\
      BBOB\_F15 & 1.7180e+2 (2.56e+1) - & \textcolor[rgb]{ .2,  .2,  .914}{3.0498e+1 (7.60e+0) +} & 5.9106e+1 (7.20e+0) \\
      BBOB\_F16 & 1.1369e+1 (2.36e+0) = & \textcolor[rgb]{ .2,  .2,  .914}{8.7828e+0 (3.10e+0) +} & 1.0439e+1 (2.28e+0) \\
      BBOB\_F17 & 4.8751e+0 (8.65e-1) - & \textcolor[rgb]{ .2,  .2,  .914}{2.0065e-2 (2.65e-2) +} & 1.7691e+0 (2.78e-1) \\
      BBOB\_F18 & 1.1648e+1 (1.51e+0) - & \textcolor[rgb]{ .2,  .2,  .914}{6.7130e-1 (7.16e-1) +} & 5.1891e+0 (8.96e-1) \\
      BBOB\_F19 & 5.6922e+0 (8.41e-1) - & \textcolor[rgb]{ .2,  .2,  .914}{2.3533e+0 (5.71e-1) +} & 3.9306e+0 (4.93e-1) \\
      BBOB\_F20 & 6.2981e+3 (2.22e+3) - & \textcolor[rgb]{ .2,  .2,  .914}{1.0869e+0 (2.45e-1) +} & 2.4717e+0 (1.59e-1) \\
      BBOB\_F21 & 2.5388e+0 (8.42e-1) - & 4.2250e-1 (4.73e-1) = & \textcolor[rgb]{ .2,  .2,  .914}{1.6986e-2 (1.60e-2)} \\
      BBOB\_F22 & 1.1268e+1 (2.30e+0) - & 6.5500e-1 (9.25e-1) - & \textcolor[rgb]{ .2,  .2,  .914}{2.8036e-1 (1.94e-1)} \\
      BBOB\_F23 & 1.7833e+0 (4.36e-1) = & \textcolor[rgb]{ .2,  .2,  .914}{1.6434e+0 (3.54e-1) =} & 1.6912e+0 (3.56e-1) \\
      BBOB\_F24 & 2.5074e+4 (1.86e+4) - & \textcolor[rgb]{ .2,  .2,  .914}{4.1991e+1 (1.07e+1) +} & 6.7391e+1 (9.38e+0) \\
      \midrule
      +/-/= & 1/21/2 & 20/2/2 &  \\
      \bottomrule
      \end{tabular}%
    \label{tabresD10}%
\end{table*}%

Traditional DE serves as a baseline and performs well on certain problems, particularly those where DDPG\_DE\_F and DQN\_DE\_MS do not show significant improvements. However, it is outperformed by DQN\_DE\_MS on the majority of the problems, underscoring the potential benefits of integrating reinforcement learning techniques into evolutionary algorithms.

DDPG\_DE\_F shows significantly worse performance on 21 out of 24 problems compared to traditional DE. This suggests that the adaptive mechanism for \( F \) may not be sufficiently robust across a wide range of optimization problems, particularly on unseen problems (e.g., BBOB\_F1, BBOB\_F5, BBOB\_F6, BBOB\_F10, BBOB\_F15, BBOB\_F16). The algorithm only outperforms DE on BBOB\_F5, where it achieves a perfect score of \( 0.0000e+0 \), indicating a potential strength in specific problem types.

DQN\_DE\_MS, demonstrates significantly better performance on 20 out of 24 problems compared to traditional DE. This highlights the effectiveness of dynamically selecting mutation strategies during the evolution process, particularly on unseen problems. The algorithm consistently achieves lower average optimization values and smaller standard deviations, indicating both better performance and higher stability.

To quantitatively evaluate the transferability of the MetaBBO algorithms, we introduce a \textbf{Transferability Index (TI)}. The TI measures the ability of an algorithm to generalize its performance from seen to unseen problems. The TI is calculated as follows:

\begin{equation}
    TI = \frac{\sum_{i \in U} \text{Perf}_i / |U| - \sum_{i \in S} \text{Perf}_i /|S| }{\sum_{i \in S} \text{Perf}_i /|S|}
\end{equation}

where \( U \) represents the set of unseen problems, \( S \) represents the set of seen problems, and \( \text{Perf}_i \) is the performance metric (the higher, the better) for problem \( i \). If TI equals to 0, it represents the performance between unseen and seen problems is the same. If TI is larger than 0, it represents the algorithm performs better on unseen problems than seen problems, indicating good transfer performance, vice versa. For example, if TI = -0.5, it can be understood that the performance of the algorithm has decreased by 50\% on unseen problems.

In this experiment, $U = \{F_1,F_5,F_6,F_{10},F_{15},F_{20}\}$, $\text{Perf}_i=-v_{avg,i}$.
For DDPG\_DE\_F: $TI_{\text{DDPG\_DE\_F}} = -9.76E-01$. For DQN\_DE\_MS: $TI_{\text{DQN\_DE\_MS}} = 1.42E+00$.
Higher TI value of DQN\_DE\_MS indicates its strong transferability, as the algorithm maintains robust performance on unseen problems, suggesting that the dynamic selection of mutation strategies generalizes well across different problem types. Lower TI value of DDPG\_DE\_F suggests poor transferability, as the algorithm struggles to adapt to unseen problems, indicating that the adaptive mechanism for \( F \) may be overfitted to the training problems.

To verify the generalization of the MetaBBO algorithm, the trained meta-optimizer trained with problems $d_{train}$ is tested across different problem dimensions $D_{diff}$. GI is introduced to quantify this performance, which can be calculated according to \eref{eqGI}.
\begin{equation}
    GI =\frac{\sum_{j \in D_{diff}} (\frac{\sum_{i\in S}(\Delta\text{Perf}_{i,j} - \Delta\text{Perf}_{i,d_{train}})}{|S|})}{|D_{diff}|}
    \label{eqGI}
\end{equation}
where $\Delta\text{Perf}_{i,j}$ represents the algorithm performance difference on problem $i$ with $D=j$ between the MetaBBO algorithm and baseline algorithm DE. 

The minimum objective results obtained by the comparative algorithms are shown in the \tref{tabresD30} and \tref{tabresD50}, with the problem dimension is 30 and 50 respectively. Same as previously, $\text{Perf}_{i,j} = -v_{avg,i,j}$ in our implementation. The GI value of DDPG\_DE\_F is -1.11E+07 while the GI value of DQN\_DE\_MS is 9.64E+05. It is obvious that the generalizability of DQN\_DE\_MS is much better than the others.

\begin{table*}[htbp]
    \centering
    \caption{Per-instance optimization results $v_{avg}(v_{std})$ of testing instances on BBOB2009 with $D=30$, where the symbols "-","+", and "=" separately indicate whether the given algorithm performs significantly worse, significantly better, or equal to traditional DE.}
      \begin{tabular}{cccc}
      \toprule
      \multicolumn{1}{c}{\textbf{Problem}} & \multicolumn{1}{c}{\textbf{DDPG\_DE\_F}} & \multicolumn{1}{c}{\textbf{DQN\_DE\_MS}} & \multicolumn{1}{c}{\textbf{DE}} \\
      \midrule
      \multicolumn{1}{c}{BBOB\_F1} & \multicolumn{1}{c}{1.4856e+2 (1.66e+1) -} & \multicolumn{1}{c}{\textcolor[rgb]{ .2,  .4,  1}{2.8916e+0 (1.31e+0) +}} & \multicolumn{1}{c}{1.2497e+1 (3.02e+0)} \\
      \multicolumn{1}{c}{BBOB\_F2} & \multicolumn{1}{c}{3.7784e+6 (8.58e+5) -} & \multicolumn{1}{c}{\textcolor[rgb]{ .2,  .4,  1}{1.5317e+4 (1.32e+4) +}} & \multicolumn{1}{c}{6.4531e+4 (1.69e+4)} \\
      \multicolumn{1}{c}{BBOB\_F3} & \multicolumn{1}{c}{9.7747e+2 (1.04e+2) -} & \multicolumn{1}{c}{\textcolor[rgb]{ .2,  .4,  1}{1.0134e+2 (3.96e+1) +}} & \multicolumn{1}{c}{3.2576e+2 (1.77e+1)} \\
      \multicolumn{1}{c}{BBOB\_F4} & \multicolumn{1}{c}{1.8883e+3 (1.73e+2) -} & \multicolumn{1}{c}{\textcolor[rgb]{ .2,  .4,  1}{1.4633e+2 (2.91e+1) +}} & \multicolumn{1}{c}{4.8666e+2 (3.75e+1)} \\
      \multicolumn{1}{c}{BBOB\_F5} & \multicolumn{1}{c}{\textcolor[rgb]{ .2,  .4,  1}{7.3448e+1 (1.78e+1) +}} & \multicolumn{1}{c}{1.8255e+2 (2.46e+1) -} & \multicolumn{1}{c}{1.5477e+2 (1.72e+1)} \\
      \multicolumn{1}{c}{BBOB\_F6} & \multicolumn{1}{c}{2.6011e+2 (3.58e+1) -} & \multicolumn{1}{c}{\textcolor[rgb]{ .2,  .4,  1}{9.4859e+0 (3.97e+0) +}} & \multicolumn{1}{c}{3.0510e+1 (7.30e+0)} \\
      \multicolumn{1}{c}{BBOB\_F7} & \multicolumn{1}{c}{6.3815e+2 (9.09e+1) -} & \multicolumn{1}{c}{\textcolor[rgb]{ .2,  .4,  1}{1.2970e+1 (6.33e+0) +}} & \multicolumn{1}{c}{5.8412e+1 (1.22e+1)} \\
      \multicolumn{1}{c}{BBOB\_F8} & \multicolumn{1}{c}{1.9326e+5 (3.29e+4) -} & \multicolumn{1}{c}{\textcolor[rgb]{ .2,  .4,  1}{4.6590e+2 (2.52e+2) +}} & \multicolumn{1}{c}{3.6651e+3 (1.14e+3)} \\
      \multicolumn{1}{c}{BBOB\_F9} & \multicolumn{1}{c}{1.7848e+5 (4.04e+4) -} & \multicolumn{1}{c}{\textcolor[rgb]{ .2,  .4,  1}{3.0334e+2 (1.28e+2) +}} & \multicolumn{1}{c}{3.2426e+3 (1.12e+3)} \\
      \multicolumn{1}{c}{BBOB\_F10} & \multicolumn{1}{c}{9.7695e+5 (2.22e+5) -} & \multicolumn{1}{c}{\textcolor[rgb]{ .2,  .4,  1}{1.5122e+4 (9.89e+3) +}} & \multicolumn{1}{c}{1.8616e+5 (4.73e+4)} \\
      \multicolumn{1}{c}{BBOB\_F11} & \multicolumn{1}{c}{4.3098e+2 (1.32e+2) -} & \multicolumn{1}{c}{\textcolor[rgb]{ .2,  .4,  1}{2.8649e+1 (7.79e+0) +}} & \multicolumn{1}{c}{1.4908e+2 (3.94e+1)} \\
      \multicolumn{1}{c}{BBOB\_F12} & \multicolumn{1}{c}{1.6429e+8 (1.70e+7) -} & \multicolumn{1}{c}{\textcolor[rgb]{ .2,  .4,  1}{2.2012e+6 (9.85e+5) +}} & \multicolumn{1}{c}{1.5481e+7 (4.28e+6)} \\
      \multicolumn{1}{c}{BBOB\_F13} & \multicolumn{1}{c}{2.2518e+3 (1.14e+2) -} & \multicolumn{1}{c}{\textcolor[rgb]{ .2,  .4,  1}{3.2228e+2 (6.55e+1) +}} & \multicolumn{1}{c}{6.3517e+2 (9.13e+1)} \\
      \multicolumn{1}{c}{BBOB\_F14} & \multicolumn{1}{c}{3.1361e+1 (6.46e+0) -} & \multicolumn{1}{c}{\textcolor[rgb]{ .2,  .4,  1}{1.2206e+0 (5.26e-1) +}} & \multicolumn{1}{c}{5.0596e+0 (1.05e+0)} \\
      \multicolumn{1}{c}{BBOB\_F15} & \multicolumn{1}{c}{9.4313e+2 (8.99e+1) -} & \multicolumn{1}{c}{\textcolor[rgb]{ .2,  .4,  1}{2.0570e+2 (2.53e+1) +}} & \multicolumn{1}{c}{3.3277e+2 (2.24e+1)} \\
      \multicolumn{1}{c}{BBOB\_F16} & \multicolumn{1}{c}{2.8892e+1 (3.23e+0) =} & \multicolumn{1}{c}{\textcolor[rgb]{ .2,  .4,  1}{2.8777e+1 (3.25e+0) =}} & \multicolumn{1}{c}{2.9656e+1 (2.29e+0)} \\
      \multicolumn{1}{c}{BBOB\_F17} & \multicolumn{1}{c}{1.0951e+1 (9.44e-1) -} & \multicolumn{1}{c}{\textcolor[rgb]{ .2,  .4,  1}{9.9083e-1 (2.91e-1) +}} & \multicolumn{1}{c}{4.4201e+0 (6.08e-1)} \\
      \multicolumn{1}{c}{BBOB\_F18} & \multicolumn{1}{c}{3.7110e+1 (1.03e+1) -} & \multicolumn{1}{c}{\textcolor[rgb]{ .2,  .4,  1}{3.9615e+0 (1.28e+0) +}} & \multicolumn{1}{c}{1.6794e+1 (2.35e+0)} \\
      \multicolumn{1}{c}{BBOB\_F19} & \multicolumn{1}{c}{2.5530e+1 (2.86e+0) -} & \multicolumn{1}{c}{\textcolor[rgb]{ .2,  .4,  1}{5.8475e+0 (3.26e-1) +}} & \multicolumn{1}{c}{8.2692e+0 (6.90e-1)} \\
      \multicolumn{1}{c}{BBOB\_F20} & \multicolumn{1}{c}{8.7566e+4 (1.50e+4) -} & \multicolumn{1}{c}{\textcolor[rgb]{ .2,  .4,  1}{2.0938e+0 (3.18e-1) +}} & \multicolumn{1}{c}{1.6266e+2 (2.04e+2)} \\
      \multicolumn{1}{c}{BBOB\_F21} & \multicolumn{1}{c}{6.4766e+1 (2.43e+1) -} & \multicolumn{1}{c}{\textcolor[rgb]{ .2,  .4,  1}{2.2345e+0 (2.75e-1) +}} & \multicolumn{1}{c}{8.6009e+0 (4.51e+0)} \\
      \multicolumn{1}{c}{BBOB\_F22} & \multicolumn{1}{c}{2.1115e+1 (9.04e-1) -} & \multicolumn{1}{c}{\textcolor[rgb]{ .2,  .4,  1}{5.6513e+0 (5.58e+0) +}} & \multicolumn{1}{c}{7.7049e+0 (1.42e+0)} \\
      \multicolumn{1}{c}{BBOB\_F23} & \multicolumn{1}{c}{\textcolor[rgb]{ .2,  .4,  1}{3.4090e+0 (4.66e-1) =}} & \multicolumn{1}{c}{3.5111e+0 (3.47e-1) =} & \multicolumn{1}{c}{3.5453e+0 (4.45e-1)} \\
      \multicolumn{1}{c}{BBOB\_F24} & \multicolumn{1}{c}{4.3563e+5 (7.86e+4) -} & \multicolumn{1}{c}{\textcolor[rgb]{ .2,  .4,  1}{2.9617e+2 (2.93e+1) +}} & \multicolumn{1}{c}{4.5367e+2 (3.83e+1)} \\
      \midrule
      +/-/= & 1/21/2 & 21/1/2 &  \\
      \bottomrule
      \end{tabular}%
    \label{tabresD30}%
  \end{table*}%

\begin{table*}[htbp]  
    \centering
    \caption{Per-instance optimization results $v_{avg}(v_{std})$ of testing instances on BBOB2009 with $D=50$, where the symbols "-","+", and "=" separately indicate whether the given algorithm performs significantly worse, significantly better, or equal to traditional DE.}
      \begin{tabular}{cccc}
      \toprule
      \multicolumn{1}{c}{\textbf{Problem}} & \multicolumn{1}{c}{\textbf{DDPG\_DE\_F}} & \multicolumn{1}{c}{\textbf{DQN\_DE\_MS}} & \multicolumn{1}{c}{\textbf{DE}} \\
      \midrule
      BBOB\_F1 & 2.9493e+2 (2.19e+1) - & \textcolor[rgb]{ .2,  .4,  1}{1.4009e+1 (3.33e+0) +} & 3.0368e+1 (6.07e+0) \\
 BBOB\_F2 & 1.0739e+7 (2.14e+6) - & \textcolor[rgb]{ .2,  .4,  1}{8.3561e+4 (4.05e+4) +} & 3.5021e+5 (9.27e+4) \\
 BBOB\_F3 & 1.8688e+3 (1.59e+2) - & \textcolor[rgb]{ .2,  .4,  1}{2.0600e+2 (3.99e+1) +} & 6.0944e+2 (4.74e+1) \\
 BBOB\_F4 & 4.1567e+3 (2.81e+2) - & \textcolor[rgb]{ .2,  .4,  1}{3.8459e+2 (8.48e+1) +} & 1.0059e+3 (9.27e+1) \\
 BBOB\_F5 & \textcolor[rgb]{ .2,  .4,  1}{2.7771e+2 (3.97e+1) +} & 4.2438e+2 (3.22e+1) = & 4.0907e+2 (3.28e+1) \\
 BBOB\_F6 & 4.9873e+2 (5.33e+1) - & \textcolor[rgb]{ .2,  .4,  1}{3.0486e+1 (9.02e+0) +} & 6.7320e+1 (1.49e+1) \\
 BBOB\_F7 & 1.5908e+3 (1.99e+2) - & \textcolor[rgb]{ .2,  .4,  1}{5.7913e+1 (1.63e+1) +} & 1.6753e+2 (3.98e+1) \\
 BBOB\_F8 & 4.2535e+5 (6.21e+4) - & \textcolor[rgb]{ .2,  .4,  1}{3.4036e+3 (1.16e+3) +} & 1.3195e+4 (5.29e+3) \\
 BBOB\_F9 & 4.4259e+5 (6.09e+4) - & \textcolor[rgb]{ .2,  .4,  1}{2.0229e+3 (9.82e+2) +} & 9.5801e+3 (2.43e+3) \\
 BBOB\_F10 & 3.4967e+6 (9.67e+5) - & \textcolor[rgb]{ .2,  .4,  1}{7.3919e+4 (4.51e+4) +} & 6.4970e+5 (1.44e+5) \\
 BBOB\_F11 & 6.6084e+2 (2.14e+2) - & \textcolor[rgb]{ .2,  .4,  1}{8.0151e+1 (1.75e+1) +} & 3.1075e+2 (6.81e+1) \\
 BBOB\_F12 & 4.2495e+8 (5.49e+7) - & \textcolor[rgb]{ .2,  .4,  1}{1.4171e+7 (4.78e+6) +} & 4.6352e+7 (8.86e+6) \\
 BBOB\_F13 & 3.1213e+3 (1.45e+2) - & \textcolor[rgb]{ .2,  .4,  1}{7.0745e+2 (1.02e+2) +} & 1.0445e+3 (1.45e+2) \\
 BBOB\_F14 & 6.4187e+1 (8.23e+0) - & \textcolor[rgb]{ .2,  .4,  1}{3.8913e+0 (1.13e+0) +} & 8.2897e+0 (1.61e+0) \\
 BBOB\_F15 & 1.7557e+3 (1.29e+2) - & \textcolor[rgb]{ .2,  .4,  1}{4.0753e+2 (5.33e+1) +} & 5.9851e+2 (5.11e+1) \\
 BBOB\_F16 & 3.7935e+1 (2.83e+0) = & 3.7775e+1 (1.77e+0) = & \textcolor[rgb]{ .2,  .4,  1}{3.7729e+1 (2.03e+0)} \\
 BBOB\_F17 & 1.4400e+1 (1.02e+0) - & \textcolor[rgb]{ .2,  .4,  1}{1.6707e+0 (3.89e-1) +} & 5.0942e+0 (6.58e-1) \\
 BBOB\_F18 & 6.2173e+1 (6.04e+0) - & \textcolor[rgb]{ .2,  .4,  1}{8.3993e+0 (1.96e+0) +} & 2.2766e+1 (2.17e+0) \\
 BBOB\_F19 & 3.1227e+1 (3.13e+0) - & \textcolor[rgb]{ .2,  .4,  1}{7.2894e+0 (3.43e-1) +} & 9.2564e+0 (7.24e-1) \\
 BBOB\_F20 & 1.9530e+5 (2.87e+4) - & \textcolor[rgb]{ .2,  .4,  1}{1.5674e+2 (3.20e+2) +} & 3.1563e+3 (1.68e+3) \\
 BBOB\_F21 & 7.9025e+1 (1.48e+0) - & \textcolor[rgb]{ .2,  .4,  1}{6.3021e+0 (2.11e+0) +} & 1.6489e+1 (3.39e+0) \\
 BBOB\_F22 & 7.9843e+1 (1.16e+1) - & 2.1661e+1 (1.22e+1) - & \textcolor[rgb]{ .2,  .4,  1}{9.5881e+0 (1.09e+0)} \\
 BBOB\_F23 & \textcolor[rgb]{ .2,  .4,  1}{4.4658e+0 (5.21e-1) =} & 4.6475e+0 (3.73e-1) = & 4.4804e+0 (4.93e-1) \\
 BBOB\_F24 & 9.0948e+5 (1.89e+5) - & \textcolor[rgb]{ .2,  .4,  1}{6.5014e+2 (5.38e+1) +} & 8.5853e+2 (5.45e+1) \\
      \midrule
      +/-/= & 1/21/2 & 20/1/3 &  \\
    \bottomrule  
    \end{tabular}%
    \label{tabresD50}%
  \end{table*}%

Conveniently, these tables can be output directly in the experimental mode of PlatMetaX GUI.

\section{Conclusion}
This work has presented a comprehensive MATLAB platform PlatMetaX for MetaBBO. Designed to address the increasing complexity of optimization challenges, PlatMetaX offers a unified framework for refining algorithms via meta-learning. By integrating a variety of meta-optimizers with a broad spectrum of base-optimizers, it provides a flexible solution for both single-objective and multi-objective optimization scenarios.

The workflow of PlatMetaX, which encompasses training and testing phases, has been delineated, showcasing the platform's capability to refine meta-optimizers for enhancing base-optimizer performance across diverse problem sets. The platform's design and resources, including its problem sets, optimizers, environment, and performance metrics, have been thoroughly described. Performance metrics for both base- and meta-optimizers have been discussed, underscoring the significance of evaluating these algorithms' effectiveness in improving optimization outcomes.

In summary, PlatMetaX stands as a notable advancement in the field of MetaBBO. Its flexible and comprehensive platform enables users to develop more effective optimization algorithms capable of automated algorithm design across various domains. The platform's design philosophy underscores modularity, ease of use, and the capacity to leverage the latest advancements in optimization and machine learning.

Despite the advancements offered by PlatMetaX, several limitations must be acknowledged. Firstly, the platform currently lacks a diverse set of black-box optimization task instances, which restricts its applicability to a broader range of real-world problems. Secondly, the evaluation metrics for meta-optimization are not fully comprehensive, potentially limiting the depth of algorithmic assessment. Thirdly, the implementation of generalization mechanisms remains relatively simplistic, which may hinder the platform's ability to adapt to varying problem domains. Lastly, the absence of detailed log files complicates the process of tracking and analyzing the optimization process. Future work should focus on expanding the repository of optimization tasks, refining the evaluation metrics, enhancing the generalization capabilities, and incorporating robust logging functionalities to improve the platform's overall utility and performance.


\bibliographystyle{IEEEtran}
\bibliography{cec2025ref}

\begin{appendices}
    \section{Implementation Tutorial for the Self-Developed MetaBBO Algorithm on PlatMetaX}
    This appendix provides a step-by-step guide for implementing the self-developed MetaBBO algorithm on the PlatMetaX platform. The instructions are structured to ensure clarity and reproducibility, following established conventions for meta-optimizer development.
    \begin{enumerate}
        \item First, define your MetaBBO components names. The MetaBBO components must be defined with specific names. For instance, the meta-optimizer, base-optimizer, and learned objects should be labeled as "MYMO", "MYBO", and "MYParas", respectively. These names are used to standardize file naming for consistent testing procedures.
        \item Second, define your environment in file \texttt{"MYMO\_MYBO\_MYParas\_Environment.m"}. In addition to the mandatory properties \textbf{State} and \textbf{IsDone}, other necessary properties of the environment must be specified. The feature vector type and size, which the meta-optimizer processes, are defined under \textbf{ObservationInfo}. Similarly, the parameter vector type and size, output by the meta-optimizer for the base-optimizer, are defined under \textbf{ActionInfo}. The initial conditions are set using the \texttt{reset} method, while the learning process of the meta-optimizer is governed by the \texttt{step} method.
        \item Third, define your base-optimizer in file \texttt{"MYMO\_MYBO\_MYParas\_Baseoptimizer.m"}, where the class \texttt{MYMO\_MYBO\_MYParas\_Baseoptimizer} inherits from \texttt{BASEOPTIMIZER}. Custom properties are defined here, and the \texttt{init} method initializes both the base-optimizer properties and the population. The \texttt{update} method represents a single evolutionary generation. The learned objects of the base-optimizer are parameterized as a variable vector within this file.
        \item Then, the meta-optimizer is defined in the file \texttt{"MYMO\_MYBO\_MYParas\_Metaoptimizer.m"}. The input and output structures are predefined in \texttt{"MYMO\_MYBO\_MYParas\_Environment.m"}. The structure or learning policy of the meta-optimizer can be customized in this file. Predefined RL agents can be utilized by renaming the file accordingly, or custom meta-optimizers can be developed by modifying the provided template.
        \item Next, the meta-optimizer is trained on a selected problem set, referred to as "problemSetName". The training is initiated by executing the command "platmetax('task', @Train, 'metabboComps', 'MYMO\_MYBO\_MYParas', 'problemSet','problemSetName')" in the command line. The training problem set is partitioned using the \textbf{splitProblemSet} function. Upon completion, the trained model is automatically saved as \texttt{"MYMO\_MYBO\_MYParas\_Metaoptimizer\_finalAgent"} in the "AgentModel" folder.
        \item Finally, the trained meta-optimizer can be tested by executing the command "platmetax('task', @Test, 'metabboComps', 'MYMO\_MYBO\_MYParas', 'problemSet','problemSetName')" in the command line. This command evaluates the meta-optimizer on predefined testing problems, as defined by the \textbf{splitProblemSet} function. Alternatively, testing can be performed using the platmetax GUI, which is launched by executing "platmetax" (without parameters) in the command line or by clicking the MATLAB "run" button when the interface is focused on the 'platmetax.m' file. The "learned" algorithm label in the GUI facilitates quick identification of the optimizer. The GUI usage mirrors that of PlatEMO, with the exception that MetaBBO methodologies require prior training via the command line.
    \end{enumerate}
    
    You can also add some problem parameters in the command when training and testing. Detailed problem parameters can be found in class \texttt{Problem}.
    
    \section{Brief Introdution of PlatMetaX GUI}
\begin{figure*}
        \centering
        \includegraphics[width = 0.95\textwidth]{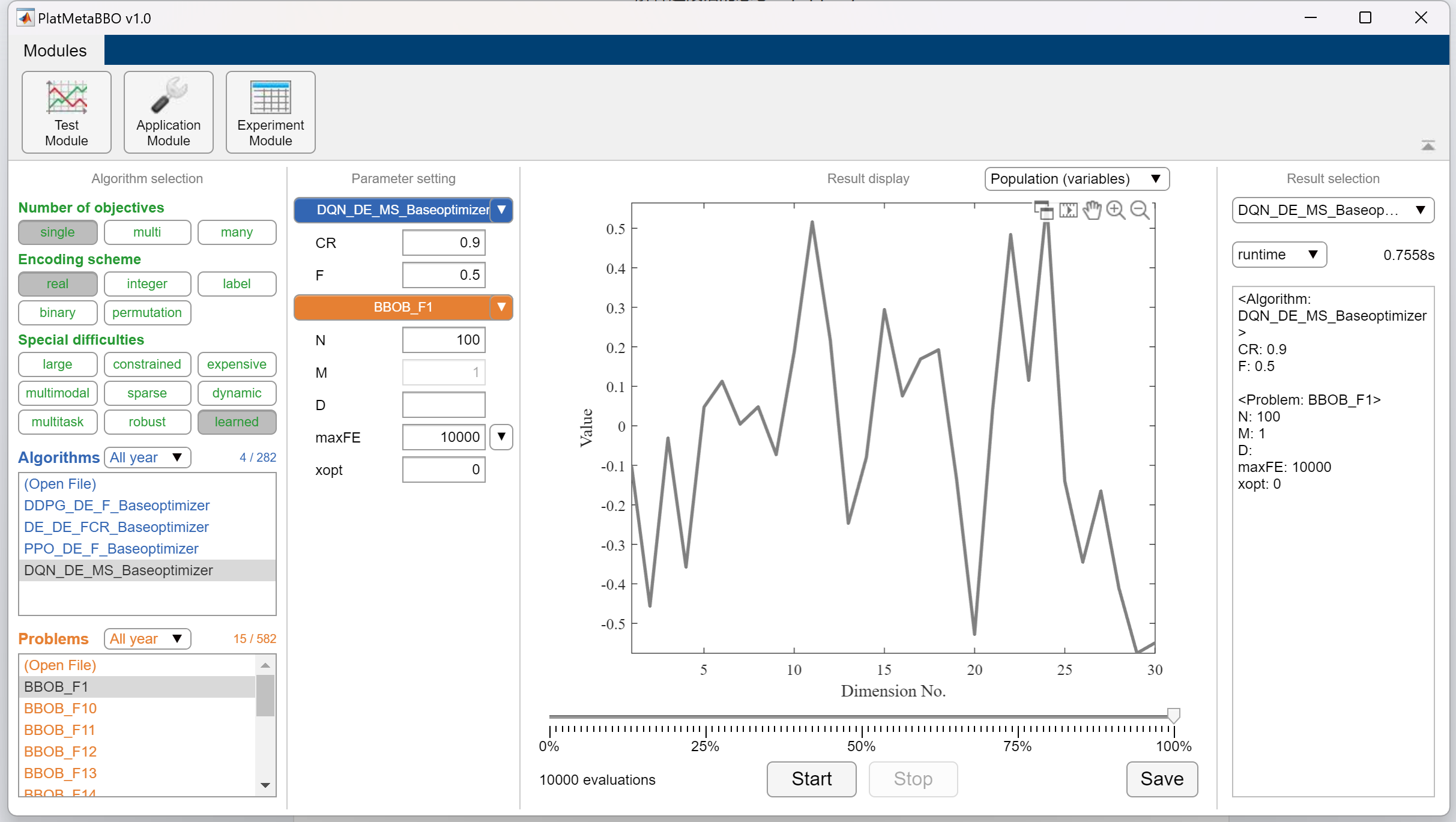}
        \caption{Test Module GUI of PlatMetaX.}
        \label{figtestmodule}
    \end{figure*}

\begin{figure*}
        \centering
        \includegraphics[width = 0.95\textwidth]{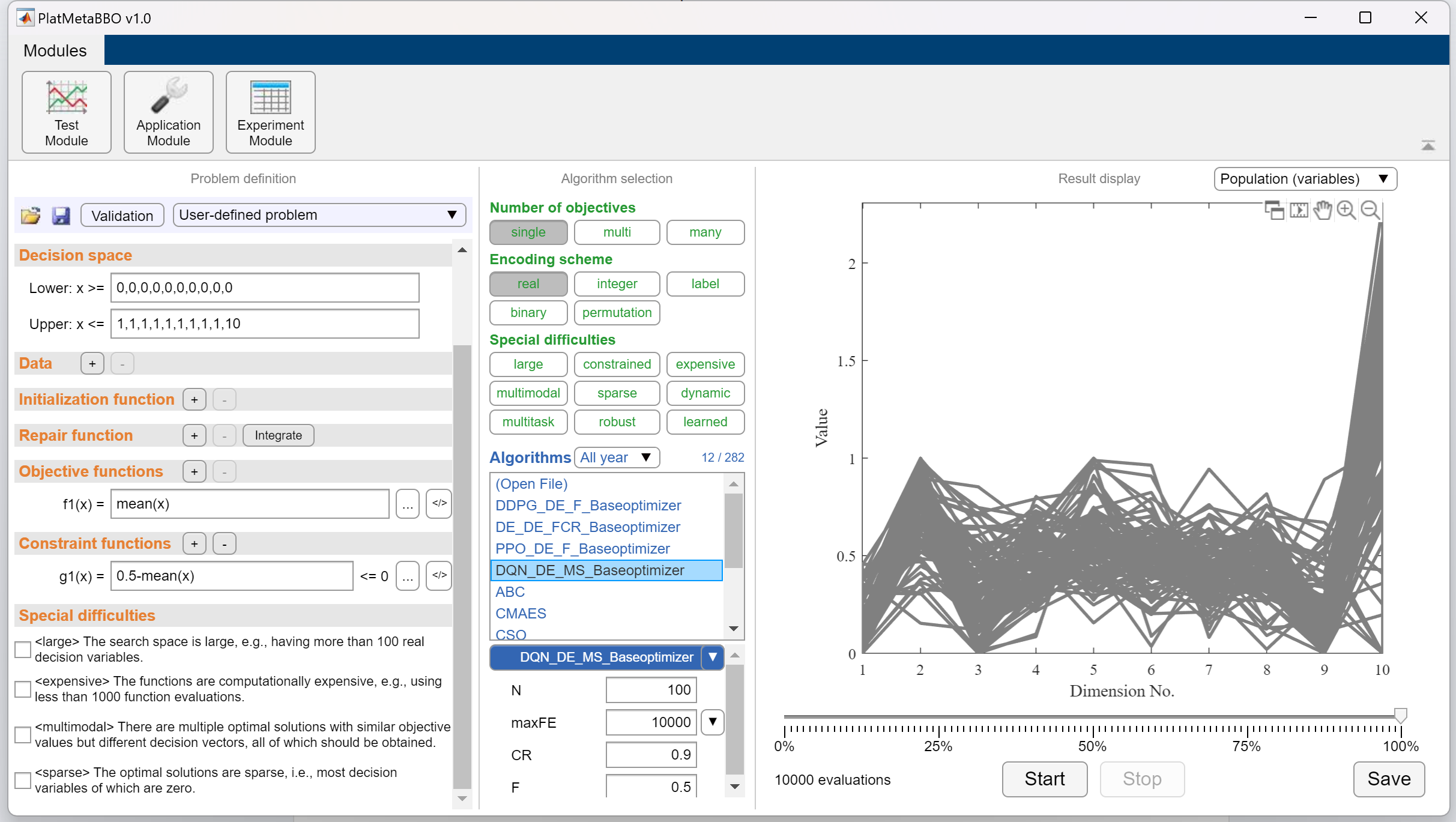}
        \caption{Application Module GUI of PlatMetaX.}
        \label{figappmodule}
    \end{figure*}

\begin{figure*}
        \centering
        \includegraphics[width = 0.95\textwidth]{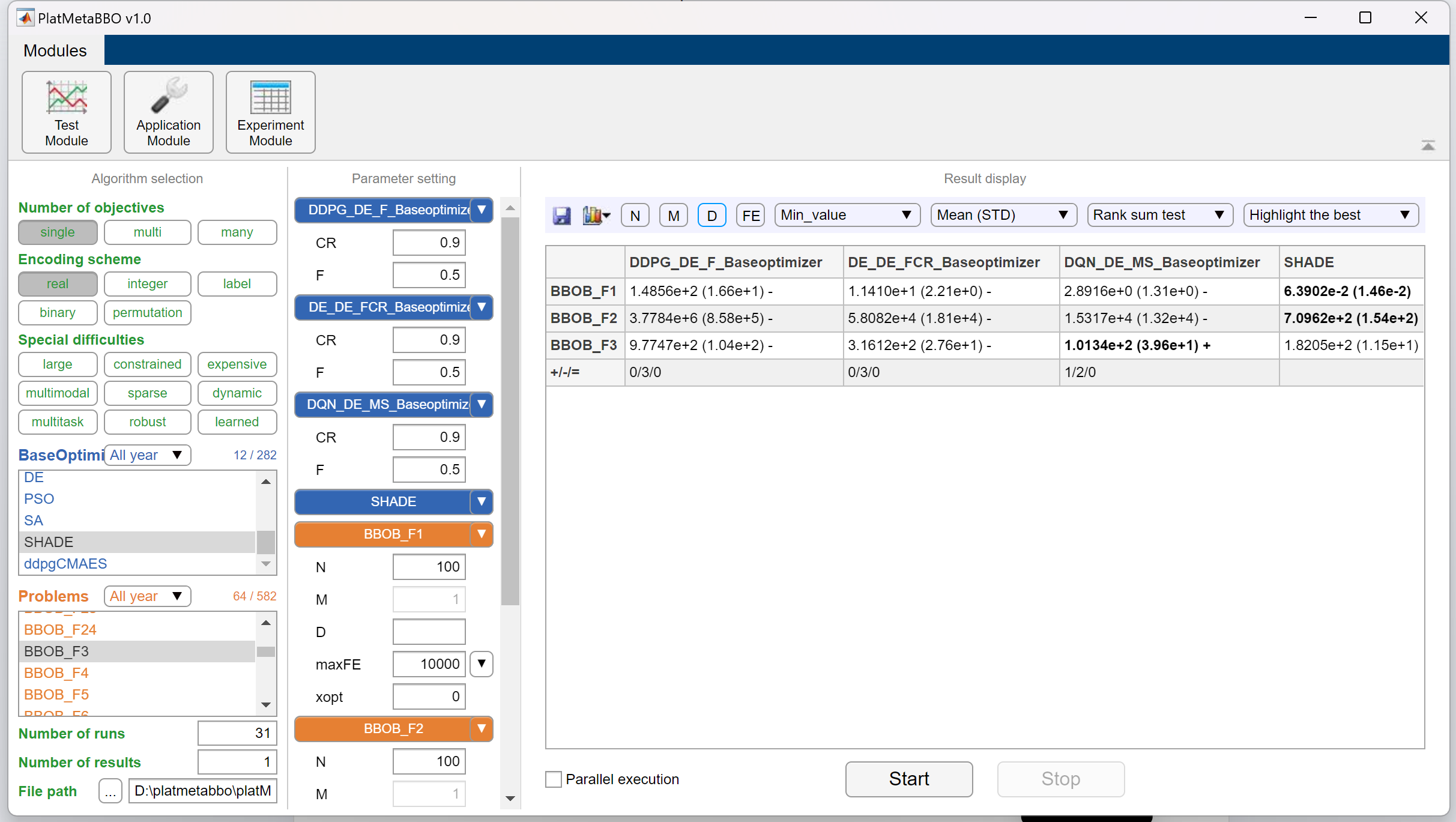}
        \caption{Experiment Module GUI of PlatMetaX.}
        \label{figexpmodule}
    \end{figure*}

    The GUI consists of three modules: test module \fref{figtestmodule}, application module \fref{figappmodule} and experiment module \fref{figexpmodule}. 
    The test module is used to test the algorithm with one selected problem and show the test results. 
    As shown in \fref{figtestmodule}, you can choose the tested algorithm and problem at the first column of the module, where you can further filter them via labels listed at the bottom part of the first column. You can set the parameters of the selected algorithm and problem at the second column. The third and fourth column present the figure results and data results separately, where the indicator can be selected in the corresponding areas. The buttons "Start", "Pause" (shown after clicking the button "Start"), and "Stop" control the test process, while the button "Save" can save the test results at where you want.
    
    As you enter application module, you can define your own problem via GUI conveniently and select any existing algorithm in PlatMetaX to solve it. Problem definition is shown at the first column, including decision space, initialization function, repari function, objective functions, constraint functions and special difficulties. The algorithm selection and its parameter setting are listed at the second column, while the third column displays the sovling process and results.
    
\begin{figure*}
        \centering
        \subfloat[Convergence on BBOB\_F1 \label{figf1}]{\includegraphics[width = 0.4\textwidth]{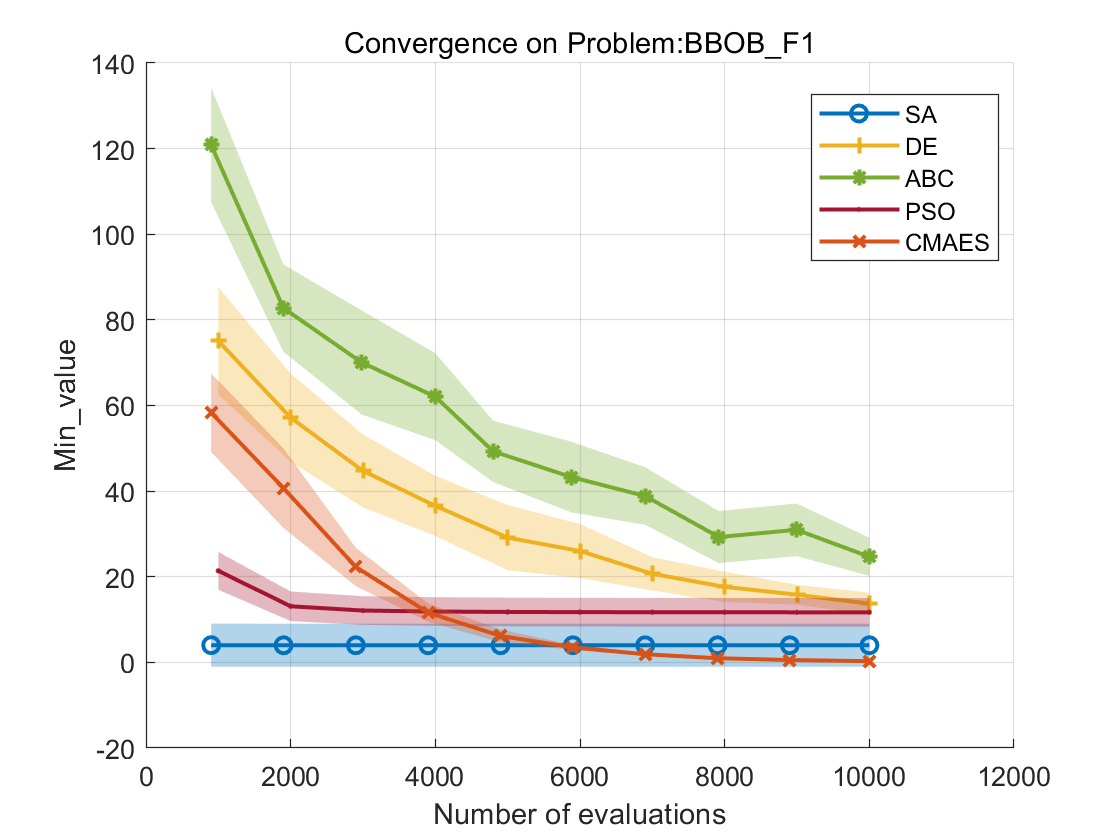}} 
        \subfloat[Convergence on BBOB\_F2\label{figf2}]{\includegraphics[width = 0.4\textwidth]{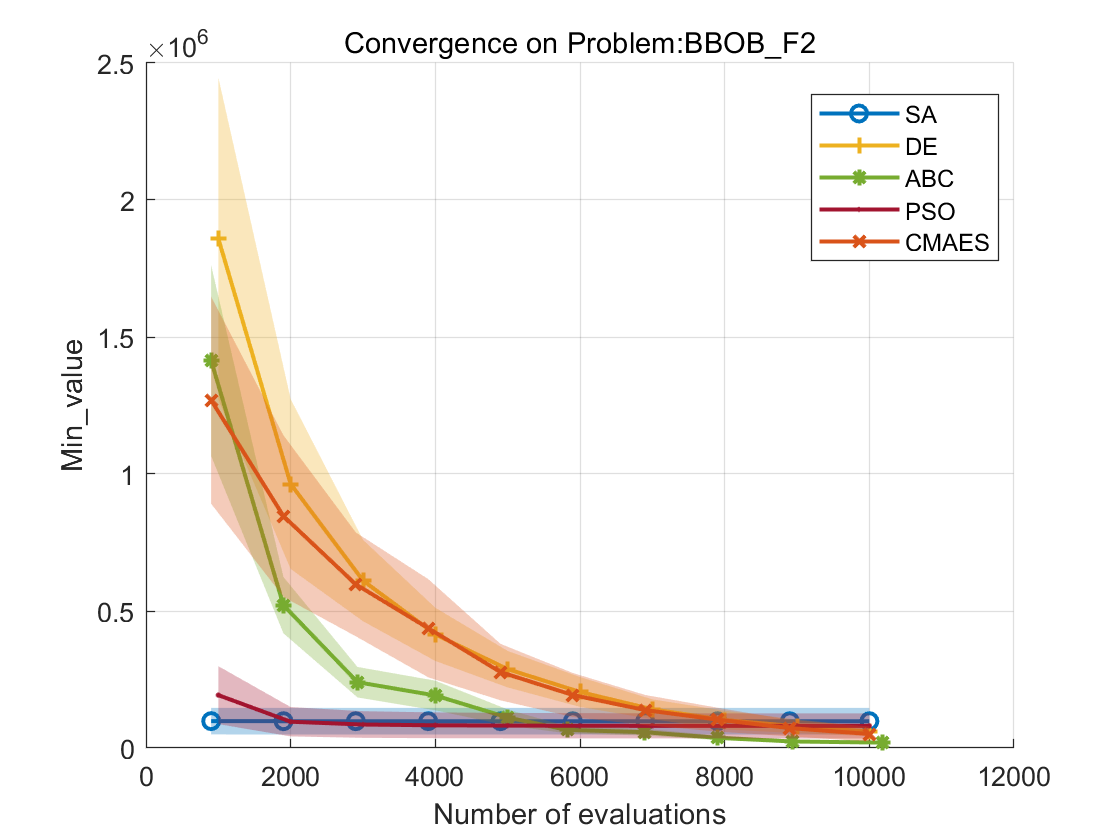}} \\
        \caption{Illustrative examples of convergence trend of different algorithms on BBOB\_F1 and BBOB\_F2 separately.}
        \label{figConvergence}
    \end{figure*}

    In the experiment module, you can test many algorithms and algorithms at once. The first column provides algorithm and problem selection, the second column provides parameter setting, and the third column shown the experimental results via table format. There are some options about experimental results that can be selected in the third column as well. It is worth noting that the experiment module is equipped with the capability to plot the convergence trends of quality indicators. An illustrative example is presented in \fref{figConvergence}. 

\end{appendices}

\end{document}